\documentclass{article}

\usepackage{arxiv}
\renewcommand{\headeright}{}
\renewcommand{\undertitle}{}
\renewcommand{\shorttitle}{Riemannian Flow Models with Reinforcement Learning for Molecular CSP}
\usepackage{etoolbox}
\makeatletter
\patchcmd{\@maketitle}{\textsc{\undertitle}\\}{}{}{\PackageWarning{manuscript}{Could not remove the preprint subtitle line}}
\makeatother
\usepackage[numbers, compress]{natbib}

\usepackage{enumitem}
\usepackage{amsmath,amssymb,amsthm}
\usepackage[most]{tcolorbox}
\usepackage{titletoc}
\usepackage{float}
\usepackage{xcolor}
\usepackage{wrapfig}

\tcbset{
  thmbox/.style={
    breakable, enhanced,
    colback=white,
    frame hidden,
    borderline west={0.8pt}{0pt}{black!35},
    boxsep=0pt,
    left=8pt, right=2pt, top=4pt, bottom=4pt,
    fonttitle=\bfseries,
    coltitle=black,
    separator sign={.\ },
    description delimiters parenthesis,
    parbox=false,
  }
}

\definecolor{xtcol}{HTML}{4561FC}
\definecolor{x0col}{HTML}{1EBA1C}
\definecolor{x1col}{HTML}{80007F}
\definecolor{vcol}{HTML}{F49D06}
\definecolor{Tx0Mcol}{HTML}{FC4545}

\usepackage{hyperref}
\hypersetup{hidelinks, pdfborder={0 0 0}}

\newtcbtheorem[number within=section]{definition}{Definition}{thmbox}{def}
\newtcbtheorem[use counter from=definition]{theorem}    {Theorem}    {thmbox}{thm}
\newtcbtheorem[use counter from=definition]{lemma}      {Lemma}      {thmbox}{lem}
\newtcbtheorem[use counter from=definition]{proposition}{Proposition}{thmbox}{prop}
\newtcbtheorem[use counter from=definition]{corollary}  {Corollary}  {thmbox}{cor}
\newtcbtheorem[use counter from=definition]{assumption} {Assumption} {thmbox}{asm}
\newtcbtheorem[use counter from=definition]{example}    {Example}    {thmbox}{ex}
\newtcbtheorem[use counter from=definition]{remark}     {Remark}     {thmbox}{rem}

\DeclareMathOperator{\Tr}{Tr}
\usepackage{graphicx}

\usepackage[utf8]{inputenc}
\usepackage[T1]{fontenc}
\usepackage{hyperref}
\usepackage{url}
\usepackage{booktabs}
\usepackage{amsfonts}
\usepackage{nicefrac}
\usepackage{microtype}

\title{Riemannian Flow Models with Reinforcement Learning for Molecular Crystal Structure Prediction}

\author{%
  \normalfont
  Thomas Egg\textsuperscript{1,2}\thanks{These authors contributed equally.},\quad
  Harry Winston Sullivan\textsuperscript{3}\footnotemark[1],\quad
  Maya M. Martirossyan\textsuperscript{1,2}, \\
  Philipp H\"ollmer\textsuperscript{1,2},\quad Cheng Zeng\textsuperscript{4,5},\quad
  Adrian Roitberg\textsuperscript{4,5},\quad Mingjie Liu\textsuperscript{4,5}, \\
  Richard Hennig\textsuperscript{5,6},\quad Sapna Sarupria\textsuperscript{7},\quad
  Ellad B. Tadmor\textsuperscript{8},\quad and Stefano Martiniani\textsuperscript{1,2,9,10} \\[5pt]
  \begin{minipage}{\dimexpr\textwidth-2\tabcolsep\relax}
  \centering\normalfont\fontsize{8}{10}\selectfont\itshape
  \textsuperscript{1}Center for Soft Matter Research, Department of Physics, New York University, New York 10003, USA\\
  \textsuperscript{2}Simons Center for Computational Physical Chemistry, Department of Chemistry, New York University, New York 10003, USA\\
  \textsuperscript{3}Department of Chemical Engineering and Materials Science, University of Minnesota, Minneapolis, MN 55455, USA\\
  \textsuperscript{4}Department of Chemistry, University of Florida, Gainesville, FL 32611, USA\\
  \textsuperscript{5}Quantum Theory Project, University of Florida, Gainesville, FL 32611, USA\\
  \textsuperscript{6}Department of Materials Science \& Engineering, University of Florida, Gainesville, FL 32611, USA\\
  \textsuperscript{7}Department of Chemistry, University of Minnesota, Minneapolis, MN 55455, USA\\
  \textsuperscript{8}Department of Aerospace Engineering and Mechanics, University of Minnesota, Minneapolis, MN 55455, USA\\
  \textsuperscript{9}Courant Institute of Mathematical Sciences, New York University, New York 10003, USA\\
  \textsuperscript{10}Center for Neural Science, New York University, New York 10003, USA
  \end{minipage}
}

\begin{document}

\date{May 6, 2026}
\maketitle

\begin{abstract} 
Crystal structure governs material properties, making crystal structure prediction (CSP) a fundamental problem in materials science. Generative models are a promising approach for solving this problem, but the prevalence of polymorphism, coupled with large unit cells and complex packing geometry, makes the molecular CSP task challenging for existing models.
To address this, we introduce Coarse-Grained Open Materials Generation (CG-OMatG), an equivariant Riemannian flow-based generative model. CG-OMatG predicts molecular crystal structures \textit{via} a coarse-grained, hierarchical representation. 
CG-OMatG treats molecules as rigid bodies---performing both inter- and intra-molecular message passing to construct a geometric representation for molecular packings---and learns to reconstruct molecule centroid positions, orientations, and lattice parameters, conditioned on chemical species and conformer geometry.
We train the model on subsets of the Open Molecular Crystals (OMC25) and Cambridge Structural Database (CSD) datasets.
Further, we fine-tune the model \textit{via} policy gradient reinforcement learning to steer the model towards generating low-energy candidate structures. 
We validate the generated structures on the CSP blind test benchmark, assessing agreement with experimentally determined crystals using COMPACK packing-similarity analysis.
CG-OMatG exhibits strong performance for generative molecular crystal structure prediction, paving the way for accelerated polymorph screening and organic solid-state materials discovery.
\end{abstract}

\section{Introduction}
\label{sec:intro}

An outstanding challenge in materials science is that of molecular (organic) crystal structure prediction (CSP), where one seeks to determine the energetically favorable structures into which organic molecules crystallize \cite{price_control_2018}. 
Unlike in atomic crystals, the weak non-bonded interactions holding molecules together in a crystal packing give rise to a proliferation of many stable energy minima, whose corresponding crystal structures can lie very close in absolute free energy \citep{nyman_static_2015} while remaining separated by large kinetic barriers. 
Molecular crystals, therefore, display a propensity for polymorphism, whereby multiple metastable crystal structures can co-exist for a given molecule, making CSP significantly more challenging than for inorganic materials. 
Different crystal polymorphs of the same compound can exhibit markedly different physical properties, only some of which may be suitable for a given application \cite{yuan_ultra-high_2014, neumann_how_2018, zhao_fat_2019, yang_deltamethrin_2020}. Correctly identifying the ground-state structure is thus critical: phase transformations away from a metastable polymorph carry nonzero probability and can compromise performance or, in pharmaceutical applications, alter drug bioavailability with direct consequences for patient safety \cite{dunitz_disappearing_1995}.
Beyond pharmaceuticals, molecular crystals find broad application across agrochemistry, food science, and electronics, making reliable polymorph prediction a problem of broad practical importance.

\begin{figure}[t]
    \centering
    \includegraphics[width=1\linewidth]{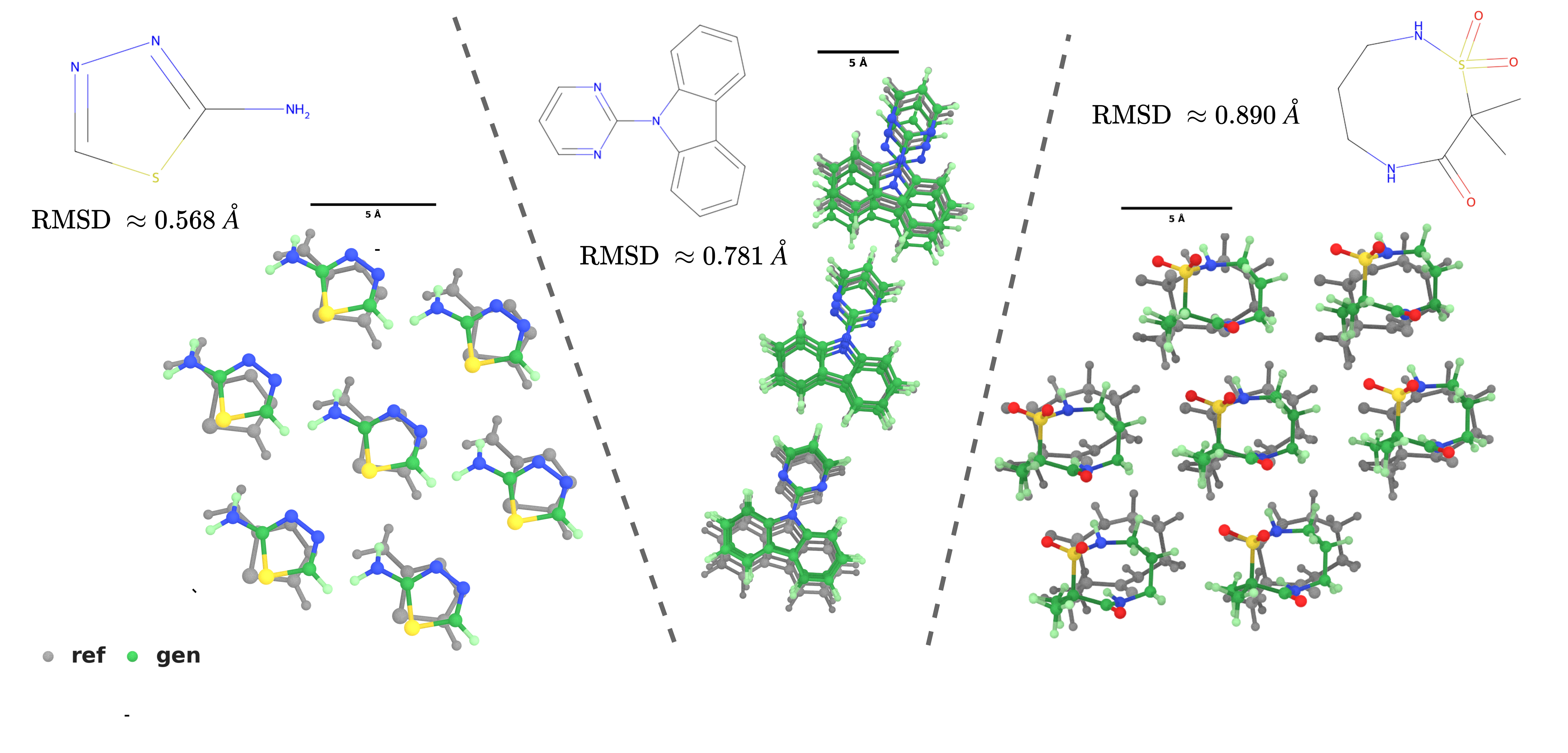}
    \caption{Comparison of generated and ground-truth crystal packings for three OMC25-MCF targets. COMPACK alignments are shown with RDKit molecular diagrams and computed $\mathrm{RMSD}_{N_{\mathrm{Matches}}}$. CSD refcodes, left to right: NIYDOO, PEMWOT, JURRET. No relaxation was performed.}
    \label{fig:omc_matches}
\end{figure}

Given the cost and difficulty of experimental crystal structure determination, substantial effort has gone into the development of computational tools for predicting these phases \textit{in silico} \cite{lommerse_test_2000, hunnisett_seventh_2024}.  
Ranking crystal structures by their energetic stability---either using quantum-mechanical calculations \textit{via} density functional theory (DFT) \cite{curtis_gator_2018} or approximating these calculations with machine learning interatomic potentials (MLIPs) \cite{gharakhanyan_fastcsp_2025,nayal_efficient_2025}---can provide insight into the available crystal structures and their relative stabilities. 
These methods typically involve many rounds of iterative optimization and expensive calculations, limiting their ability to exhaustively explore all low-energy phases.

Recent advances in MLIPs and generative models offer a more scalable path, enabling faster exploration of the energy landscape and, through guidance \cite{ho_classifier-free_2022, prakash_guided_2025, skreta_feynman-kac_2025, domingo-enrich_adjoint_2025a} and reinforcement learning  \cite{shao_deepseekmath_2024, hoellmer_open_2026}, the ability to steer generation toward desired properties. Applying these tools to molecular CSP, however, introduces distinct challenges. All-atom generative models lack explicit knowledge of the different length scales and bond types present in organic molecules,  and there is no constraint enforcing that intramolecular bonds remain coherent during or at the end of generation.
Unit cell sizes compound the problem: organic crystals are typically far larger than the inorganic structures for which most generative models have been designed and trained \cite{martirossyan_all_2025}.
Existing approaches must therefore be substantially tailored for molecular crystal structure prediction.

We present a coarse-grained generative model for molecular crystal structure prediction, following a line of work that treats molecules as rigid-body building blocks \cite{guo_assembleflow_2024, kim_mofflow_2025, zeng_molcrystalflow_2026}. By separating the strong intramolecular interactions that define molecular shape from the weak intermolecular interactions that drive the crystal packing, this approach directly addresses the multi-scale character of molecular crystals, focusing generation on the placement and orientation of rigid molecules within the unit cell. \citet{galanakis_rapid_2024} have shown that molecular centers of mass tend to occupy well-defined packing positions, suggesting that crystal packings can be learned effectively at this coarser scale. Our model learns local-coordinate representations and constrains the generative process to building block positions and orientations, encoding this multi-scale inductive bias by design.

\newpage 

\paragraph{Our contributions}

\begin{itemize}[leftmargin=*,itemsep=1pt,topsep=2pt,parsep=0pt]
    \item We introduce the \textbf{C}oarse-\textbf{G}rained \textbf{O}pen \textbf{Mat}erials \textbf{G}enerator (CG-OMatG), an equivariant generative model for molecular CSP.

    \item We mathematically formulate the Riemannian manifold of molecular crystal configurations and construct an affine-invariant geodesic unit-cell interpolation using polar decomposition on $SO(3) \times \mathrm{Sym}_3^+$, ensuring every point along the path corresponds to a nondegenerate, positive-volume unit cell---in contrast to prior parameterizations on $\mathbb{R}^{3\times 3}$. 

    \item We formulate a group-relative policy optimization (GRPO) scheme on the manifold of molecular crystal configurations and apply it to reward energetically stable packings.
\end{itemize}

\section{Related Work}

\subsection{Molecular CSP}

Predicting energetically favorable molecular crystal packings is a long-standing challenge with far-reaching implications for materials design, pharmaceutical development, and discovery of functional molecular solids \cite{price_control_2018}. 
Due to the considerable cost, effort, and time required to identify crystal structures experimentally, there is significant interest in computational approaches that can accelerate the CSP pipeline.  
Conventional methods rely on iterative rounds of expensive quantum chemical calculations \cite{price_predicting_2014, beran_modeling_2016}.
Recent work seeks to bypass expensive energy function evaluations entirely: \citet{galanakis_rapid_2024} introduced CrystalMath, an optimization procedure that optimizes the crystal structure with respect to simple order parameters and reports rapid structure prediction for systems with one or two molecules $Z' \in \{1, 2\}$ in the asymmetric unit.
Other tools like Genarris and FastCSP couple random structure generation with physical constraints or energy evaluations to predict stable configurations of close-packed molecular crystals \cite{li_genarris_2018, yang_genarris_2025, gharakhanyan_fastcsp_2025}.

\subsection{Generative Models}

Developments in machine learning have spurred the advent of generative models which accelerate crystal structure prediction by learning to sample from the distribution of known crystal structures, obtained either through experimental determination or first-principles calculations.
A wealth of models have been devised to predict inorganic condensed phases conditioned on a target chemistry \cite{xie_crystal_2022, jiao_crystal_2024, miller_flowmm_2024, zeni_generative_2025, antunes_crystal_2024, hoellmer_open_2025, veljkovic_crystalite_2026}, 
and adaptations for larger and more complex systems have followed, including models for metal-organic frameworks \cite{kim_mofflow_2025, kim_flexible_2025, simkus_mofasa_2025}.

Generative models for molecular CSP are more recent.  AssembleFlow uses inertial frames to decompose molecular $SE(3)$ transformations into separate translation and rotation flows for finite molecular clusters \cite{guo_assembleflow_2024}.
OXtal, an AlphaFold3-style diffusion model, learns crystalline packings of molecular conformers in Cartesian space, foregoing the learning of the unit-cell lattice \cite{jin_oxtal_2025}.  
Both of these approaches, however, require post-hoc Patterson analysis \cite{patterson_fourier_1934} to recover the unit cell, limiting their utility for property-based guidance based on energy calculations.

Closest to our work, MolCrystalFlow \cite{zeng_molcrystalflow_2026} and PackFlow \cite{subramanian_packflow_2026} also apply flow-based generative modeling to molecular CSP. MolCrystalFlow shares the rigid-body decomposition and Riemannian treatment of molecular degrees of freedom with CG-OMatG, but parameterizes the lattice as an unconstrained matrix in $\mathbb{R}^{3\times 3}$ and does not include RL post-training. PackFlow generates per-atom Cartesian coordinates for all heavy atoms in the unit cell without exploiting the rigid-body structure of molecular crystals, jointly sampling these with lattice parameters in Euclidean space. 
Related lattice decompositions are used by MatterGen \cite{zeni_generative_2025} and DiffCSP++ \cite{jiao_space_2024}, the latter preserving positive definiteness through diffusion in a symmetric logarithmic representation. CG-OMatG instead combines affine-invariant geodesic unit-cell interpolation with coarse-grained building blocks and GRPO on $SO(3) \times \mathrm{Sym}_3^+$.

\subsection{Reinforcement Learning}

\textit{Post-training} via reinforcement learning (RL) provides a way to align generative models with downstream objectives by optimizing neural network weights against a reward function. RL post-training has begun to show success in generative modeling for inorganic materials \cite{Karpovich2024,CrysText,plaid,park_guiding_2025, chen_accelerating_2025,Cao_2026}. \citet{hoellmer_open_2026} demonstrated the potential of RL to steer pretrained inorganic CSP models toward low-energy structures in OMatG-IRL, and \citet{subramanian_packflow_2026} pursued an analogous objective in the molecular setting in PackFlow. Both works address the central challenge of applying policy-gradient RL to ODE-based generative models, but through distinct constructions. OMatG-IRL introduces stochasticity into the ODE dynamics, yielding a surrogate SDE with tractable step-wise transition likelihoods that provide exact importance ratios and KL terms for policy-gradient updates. PackFlow retains deterministic ODE sampling and instead approximates per-sample policy scores using the flow-matching pretraining loss evaluated at a single time point, yielding surrogate importance ratios and KL regularization terms that do not correspond to exact likelihoods of the generative process. The RL setup in CG-OMatG builds on the construction in OMatG-IRL, extending it from Euclidean space to the Riemannian manifold of molecular crystal configurations, where stochastic exploration is introduced in the tangent space of this manifold, enabling policy-gradient RL with exact transition probabilities while preserving the geometry of the generative dynamics.

\section{Methods}

\subsection{Molecular Crystal Structure Prediction} \label{sec:molecularcsp}

\paragraph{Crystal Structure Prediction} The CSP task can be framed as a sampling problem targeting a conditional distribution
\begin{align}
    p \left( L,\{c^{(j)}\}_{j=1}^N \mid \mathcal{A} \right) = p(y \mid \mathcal{A}), \label{eq:csptask}
\end{align}
where $L \in GL^+(3,\mathbb{R})$ is a row-major matrix of lattice vectors, $c^{(j)} \in \mathbb{R}^3$ is the Cartesian position of atom $j$, and $\mathcal{A} := \{a^{(j)}\}_{j=1}^N$ where $a^{(j)} \in \{0,1\}^T$ is a one-hot vector encoding its atomic type. The molecular crystal definition is formulated rigorously in Definition~\ref{def:crystal}. In principle, this distribution is induced by the laws of quantum mechanics and thermodynamics (the latter only for nonzero temperatures). For the purposes of generative model training we use a dataset $\mathcal{D}$ of energetically stable or experimentally realizable crystals as a proxy for Equation~\ref{eq:csptask}.

\paragraph{Factorization of the Joint Distribution} In this work we exploit the hierarchical structure of molecular crystals to sample from a factorization of Equation~\ref{eq:csptask}. Splitting the joint distribution into intra- and inter-molecular factors lets us enforce strict molecular validity without sacrificing probabilistic consistency. To do so we consider a coarse-graining map (Definition~\ref{def:cgmap}) that removes all intramolecular degrees of freedom, replacing each molecule $i$ with a centroid $q^{(i)} \in \mathbb{R}^3$ and an orientation $Q^{(i)} \in SO(3)$ relative to a set of canonical coordinates $\{\tilde c^{(j)}\}_{j \in S_i}$. Partitioning the atoms into $M$ disjoint molecular subsets $\{S_i\}_{i=1}^M$ and applying the coarse-graining map to each allows us to rewrite the target in Equation~\ref{eq:csptask}
\begin{align}
    p\left(L, \{c^{(j)}\}_{j=1}^N \middle | \mathcal{A}\right)
    = p\left(L, \{q^{(i)}, Q^{(i)}\}_{i=1}^M \,\middle|\, C, \mathcal{A}\right)\, p(C| \mathcal{A}),
\end{align}
where $C := \big(\{\tilde c^{(j)}\}_{j \in S_1}, \dots, \{\tilde c^{(j)}\}_{j \in S_M}\big)$ collects the local atomic coordinates of each molecule in its frame. The first factor is the inter-molecular distribution over cell and rigid-body placements; the second is the intra-molecular conformer prior. 

We further assume local atomic coordinates are mutually independent across molecules,
\begin{align}
    p(C \mid \mathcal{A}) \approx \prod_{i=1}^{M} p\left(\{\tilde{c}^{(j)}\}_{j\in S_i} \,\middle|\, \{a^{(j)}\}_{j\in S_i}\right), \label{eq:factorizeprior}
\end{align}
which is reasonable when molecules are sufficiently rigid and only weakly perturbed by their environment. 
Additionally, we assume each factor is concentrated around an \emph{a priori} known conformer, so that $C$ may be treated as fixed (rigid-body assumption). Together, these reduce the learning problem to the inter-molecular factor alone. Both assumptions can fail for flexible molecules, and we leave their relaxation to future work. In the absence of the true conformer, one must estimate it by another method before applying the current iteration of CG-OMatG.

We parameterize molecular centroid translations in fractional coordinates $f^{(i)} = \operatorname{wrap}(q^{(i)}L^{-1}) \in \mathbb{T}^3$, which simplifies the implementation of periodic boundary conditions \cite{miller_flowmm_2024}. 
The lattice matrix $L$ itself also requires a parameterization suitable for Riemannian flow matching. Naively treating $L$ as an element of $\mathbb{R}^{3\times 3}$ does not guarantee that intermediate points along a flow remain valid unit cells. To address this, we decompose $L$ via the polar decomposition \cite{horn_matrix_2012}. Considering, for simplicity, the column-major representation of $L$ which we write $L'=L^\top$, this factors $L' = UP$ uniquely into a symmetric positive-definite $P = (L'^{\top} L')^{1/2}$ and an orthogonal $U = L'P^{-1}$. Taking determinants gives $\det L' = \det U \det P$; since $\det P > 0$ and $\det L' > 0$ (as $L' \in GL^+(3,\mathbb{R})$), we have $\det U = 1$, so $U \in SO(3)$. This gives the manifold of unit cells as $
    GL^+(3,\mathbb{R}) \cong SO(3) \times \mathrm{Sym}_3^+,
$
where $\mathrm{Sym}_3^+$ is the set of symmetric positive-definite $3\times 3$ matrices. This choice ensures that at all times during flow the unit cell is nondegenerate and has positive volume; we comment briefly on this point in Appendix~\ref{app:euc_unitcell}. For $L'=UP$, the metric tensor is $G=L'^{\top}L'=P^2$. We retain $U$ because the cell and molecular orientations share a Cartesian frame, and we do not pre-rotate the OMC or CSD data during preprocessing. Together these identities allow us to write the target probability data distribution as
\begin{align}
    \varphi(x) := p\left(U, P, \{f^{(i)}, Q^{(i)}\}_{i=1}^{M} \,\middle|\, C, \mathcal{A}\right).
\end{align}

\paragraph{Molecular Crystal Manifold} The variable $x:=(U, P, \{f^{(i)}, Q^{(i)}\}_{i=1}^M)$ lives on a partially curved product space rather than Euclidean space. 
To apply Riemannian flow matching, we collect these variables into a single product manifold  $\mathcal{M}$ defined as
\begin{align}
    \mathcal{M} := SO(3)\times \mathrm{Sym}_3^+ \times \left(\mathbb{T}^3 \times SO(3)\right)^M.
\end{align}
Each point $x$ specifies a molecular crystal configuration.
The complete definition of the manifold along with its Riemannian metric is provided in Appendix~\ref{app:mol_cryst_manifold}.
By summing the metric on each sub-manifold, $\mathcal{M}$ is trivially a Riemannian manifold (see \cite[Eq.~3.3]{lee_introduction_2018} and \cite[Examples 1.8 and 13.2]{lee_introduction_2012}). 
The induced distance, logarithm, and exponential maps are given in Appendix~\ref{sec:explogdist}; they define the closed-form geodesics used to construct the conditional velocity field in Section~\ref{sec:geometric_flow}.

\subsection{Learning Crystal Packings with Riemannian Flow Models} \label{sec:geometric_flow}

\paragraph{Geometric Flow Modeling} Assuming access to samples $x_1$ from some unknown data distribution $\varphi: \mathcal{M} \to \mathbb{R}_{\geq 0}$ along with an easy-to-sample prior $p_0: \mathcal{M} \to \mathbb{R}_{\geq 0}$, the goal is to learn a bijection $\Psi: \mathcal{M}\to\mathcal{M}$ which pushes $p_0$ forward to closely approximate $\varphi$. 
We learn this bijection by borrowing ideas from dynamical measure transport \cite{liu_flow_2022a, albergo_building_2023, lipman_flow_2023}.
Specifically, we aim to parameterize a time-dependent velocity field $u_t:[0,1]\times\mathcal{M}\to T_x\mathcal{M}$ in the manifold ODE
\begin{align}
    \tfrac{d}{dt}\psi_t(x) = u_t(\psi_t(x)), \qquad \psi_0(x)=x.
    \label{eq:mode}
\end{align}
with solution $\psi_t$. The flow map induces a family of pushed forward densities of the form
\begin{align}
    \log p_t(x)=\log p_0\big(\psi_t^{-1}(x)\big)-\int_{0}^{t}\operatorname{div}_g\big(u_s(x_s)\big)ds. 
\end{align}
As the solution to the ODE is deterministic and may be time-reversed, we know it is invertible. This lets us define the bijection $\Psi$ as the $t=1$ solution of this ODE, \textit{i.e.}, $\Psi := \psi_1$. The goal then is to approximate $u_t$ by some neural network $b^\theta_t:\mathcal{M}\to T_x\mathcal{M}$, which in turn induces an approximate bijection which can be used for generating molecular crystal configurations. 

\paragraph{Riemannian Conditional Flow Matching} Chen and Lipman \cite{chen_flow_2024} show that the minimizer of the Riemannian conditional flow matching (RCFM) objective provides a training target for the model velocity $b_t^\theta$ such that the pushforward satisfies $p_1 \approx \varphi$ when optimized. 
The RCFM loss is
\begin{align}
\mathcal{L}[b_t^\theta]
=\mathbb{E}\left[\big\|b_t^\theta(x)-u_t(x\mid x_1)\big\|_g^2\right],
\label{eq:loss}
\end{align}
where the expectation is taken over $
t \sim U([0,1]), x_1 \sim \varphi(x_1),$ and $ x \sim p_t(x\mid x_1).
$
The function $p_t(\cdot\mid x_1):\mathcal{M}\to\mathbb{R}_{\geq0}$ denotes a conditional probability path satisfying the boundary conditions
$
p_1(x\mid x_1) = \delta_{x_1}(x)$ and $
p_0(x\mid x_1) = p_0(x),
$
meaning that at $t=1$ it is tightly distributed about the conditioning point, while at $t=0$ it reproduces the easy-to-sample prior. 
This density is induced by a conditional velocity $u_t(x\mid x_1)$ whose ODE solution is $\psi_t(x\mid x_1)$, with initial condition
$
\psi_0(x\mid x_1)=x,
$
generates the conditional density via the push-forward. 

\paragraph{Parameterization of Conditional Velocity} We parametrize the conditional velocity using geodesics on $\mathcal{M}$ defined through the Riemannian logarithm and exponential maps. 
Specifically we set
\begin{align}
    x_t := \psi_t(x_0|x_1) =  \exp^\mathcal{M}_{x_0}\left(t\log^\mathcal{M}_{x_0}(x_1)\right).
\end{align}
where $x_t$ is just shorthand for the conditional ODE solution $\psi_t(x_0|x_1)$ as visualized in Figure~\ref{fig:manifold} in Appendix section \ref{app:mol_cryst_manifold}. Since $\mathcal{M}$ is a product manifold, this path is obtained by evolving each component along its corresponding geodesic. For the fractional coordinates this gives the minimum-image straight line on the torus \cite{hoellmer_open_2025},
\begin{align}
    f_t
    =
    \psi_t^{\mathbb{T}^3}(f_0\mid f_1)
    =
    f_0+t(f_1-f_0-n^\star)\ \mathrm{mod}\ \mathbb{Z}^3,
\end{align}
where $n^\star\in\arg\min_{n\in\mathbb Z^3}\|(f_1-f_0)-n\|$ is the integer translation that minimizes the distance between $f_0$ and $f_1$ under periodic boundary conditions. For rotations, the same geodesic applies to both the molecular orientation $Q_t$ and the cell orientation $U_t$. Writing generically $R_t\in SO(3)$ with endpoints $R_0$ and $R_1$, and defining
\(
    \Omega
    =
    \log_{R_0}^{SO(3)}(R_1),
\) and \(
    \theta
    =
    (-\tfrac12\operatorname{Tr}\left((R_0^\top\Omega)^2\right))^{0.5},
\)
the geodesic is evaluated using the Rodrigues formula \cite{rodrigues_lois_nodate}
\begin{align}
     R_t
    =
    \psi_t^{SO(3)}(R_0\mid R_1)
    =
    R_0\left(
    I
    + \frac{\sin(t\theta)}{\theta}(R_0^\top\Omega)
    + 2\frac{\sin^2(t\theta/2)}{\theta^2}(R_0^\top\Omega)^2
    \right).
\end{align}
Finally, the symmetric positive-definite cell component follows the affine-invariant geodesic \cite{pennec_riemannian_nodate}
\begin{align}
    P_t
    =
    \psi_t^{\mathrm{Sym}_3^+}(P_0\mid P_1)
    =
    P_0^{\frac12}\exp\left(
        t\,\log\left(P_0^{-\frac12}P_1P_0^{-\frac12}\right)
    \right)P_0^{\frac12},
\end{align}
where the matrix logarithm, matrix exponential, and principal square root are each evaluated by diagonalizing the argument, applying the corresponding scalar function to the eigenvalues, and reconstructing the matrix from the resulting spectrum.

\subsection{Constraints on the Velocity $b^\theta_t$}

\paragraph{Tangency Constraints} A standard flow model parametrizes a map $\mathbb{R}^d\to\mathbb{R}^d$, but on a manifold the velocity must satisfy $b_t^\theta(x)\in T_x\mathcal{M}$. Two common methods to enforce this are projecting an ambient prediction onto the tangent space via $P_x:\mathbb{R}^{\mathrm{emb}(\mathcal{M})}\to T_x\mathcal{M}$~\cite{gemici_normalizing_2016, lou_neural_2020, mathieu_riemannian_2020, chen_flow_2024}, or predicting an endpoint $\hat{x}_1\in\mathcal{M}$ and recovering the velocity as $b_t^\theta(x_t)=\log_{x_t}^{\mathcal{M}}(\hat{x}_1)$~\cite{yim_se3_2023, kim_mofflow_2025, zeng_molcrystalflow_2026}. Following the generator-based construction of Falorsi and Forr\'e~\cite[Appendix~B.2]{falorsi_neural_2020}, we use a third parameterization based on the Lie group structure: the network outputs an unconstrained $\omega\in\mathbb{R}^{\dim\mathfrak{g}}$, which is hat-mapped to the Lie algebra $\widehat{\omega}\in\mathfrak{g}$ and left-translated to give $b_t^\theta(x_t) = x_t\,\widehat{\omega}$. Since $T_x\mathcal{M}=x\,\mathfrak{g}$ for any matrix Lie group, tangency holds by construction. Our model uses all three parameterizations. The cell rotation head takes the Lie-algebra route, outputting $\omega\in\mathbb{R}^3$ and hat-mapping to $\widehat{\omega}\in\mathfrak{so}(3)$ to obtain a velocity in $T_{U_t}\mathrm{SO}(3)$. The per-molecule orientation head uses the log-map, the lattice shape head uses projection via Voigt-vector readout, and the centroid head is Euclidean. Full per-head details are in Appendix~\ref{app:architecture}.

\paragraph{Symmetry Constraints} Crystal structures, like many physical systems, are known to exhibit symmetries corresponding to conservation laws \cite{noether_invariante_1918}. It is well established that accounting for symmetries leads to higher-quality samples and networks that generalize better \cite{lecun_backpropagation_1989, schutt_equivariant_2021, bose_equivariant_2022, weiler2023EquivariantAndCoordinateIndependentCNNs,hou_score_2025,domina_how_2026}. Molecular crystals exhibit a very rich symmetry structure: translations, rotations, lattice basis changes, PCA sign ambiguity \cite{li_closer_2021}, conformer point group operations, space group operations, and permutations of both atom and molecule labels. A complete mathematical treatment is deferred to Appendix~\ref{app:symmetries}; here we focus on the subset of symmetries that our generative framework must actively handle. Abstractly, a symmetry group $G$ acts on the manifold $\mathcal{M}$. For $x\in\mathcal{M}$, its orbit is $Gx = \{gx \mid g\in G\}$, and the quotient $\mathcal{M}/G$ identifies configurations that lie in the same orbit. It would therefore be natural to define the velocity directly on the quotient, with $b^\theta_t([x])\in T_{[x]}(\mathcal{M}/G)$ \cite{zhou_rethinking_2026}. This strategy has been applied to space-group-constrained crystal generation \cite{jiao_space_2024}, where the Wyckoff position provides a natural representative of each orbit, and to pose prediction \cite{levy-jurgenson_manifold_2026} by selecting the rotation closest to the identity as a canonical representative. In both cases, the symmetry is quotiented out by choosing a representative $x_{\text{rep}}([x]) \in [x]\subset\mathcal{M}$ prior to training. Such canonicalization would require choosing an orbit representative before training; we do not impose that preprocessing convention here.

\paragraph{Equivariant Networks} Although we could work on the quotient space so that each state corresponds to a unique molecular crystal, we instead adopt an equivariant modeling approach. Kohler \textit{et al.} \cite{kohler_equivariant_2020} proved that the pushforward of a $G$-equivariant $\psi_t$ produces a $G$-invariant density as long as the base density is at least $G$-invariant. Corresponding extensions to Riemannian flow models have been proven as well \cite{katsman_equivariant_2022}. Their theorems state that the velocity network $b^\theta_t$ must be \emph{equivariant}. Letting $\Phi_g(x) = gx $ denote the left group action, we require for each symmetry $g\in G$ and for each point on the manifold $x\in\mathcal{M}$
\begin{align}
    b^\theta_t(\Phi_g(x)) =(d\Phi_g)_x(b^\theta_t(x)) \in T_{\Phi_g(x)}\mathcal{M}
\end{align}
where the differential acts as $(d\Phi_g)_x:T_x\mathcal{M}\to T_{\Phi_g(x)}\mathcal{M}$ and represents the action of the symmetry on the velocity. This is the differential geometric version of the rather intuitive statement: when the system is rotated, the velocity vectors rotate with it. In Appendix \ref{app:symmetries} we describe the differential for the rotation and translation groups ($SO(3)$ and $\mathbb{R}^3$ respectively) acting on configurations $x\in\mathcal{M}$, and we further apply an equivariant neural network to enforce this symmetry directly \cite{geiger_e3nn_2022}. A full prescription of the architecture with numerical estimates of the equivariance error \cite{domina_how_2026} is reported in Appendix \ref{app:architecture}. 

We also apply a new type of data augmentation that handles both the frame ambiguity and the point group ambiguity of the CG pose assigned to a given molecule. We perturb atomic positions with small isotropic noise ($\approx 0.01\;\text{\AA}$), apply the coarse-graining map in Definition~\ref{def:cgmap} to obtain a unique orientation from the noisy positions, and then express the original un-noised positions in the resulting orientation to obtain local coordinates. This procedure spans the set of possible poses that can be assigned to a molecule with non-trivial symmetries, allowing the training procedure to be robust to the inherent ambiguity in orientation assignment and removing the need for canonicalization present in other point cloud pose estimation schemes~\cite{levy-jurgenson_manifold_2026}. Rather, we simply train on all equivalent orientations.

\subsection{Reinforcement Learning on the Molecular Crystal Manifold} \label{sec:rl_main}

Flow models can be post-trained with Flow-GRPO \cite{liu_flow-grpo_2025}, which converts a flow ODE into a marginally equivalent SDE. However, its derivation requires a Gaussian base distribution and a linear interpolant, neither of which is available for our manifold-valued flow. We therefore cannot apply Flow-GRPO directly and instead build on the more general surrogate-SDE framework of \citet{hoellmer_open_2026}, extending it to dynamics on $\mathcal{M}$. The resulting log-likelihood factorizes into a tangent-space Gaussian and a parameter-independent Jacobian that cancels from the importance ratio and KL term. Alternative RL formulations for materials generation include Reinforce Adjoint Matching, used by OMatG-flash to post-train flow maps~\cite{egg_omatg_flash_2026}.

\paragraph{Promotion of the ODE to an SDE} To enable exploration during RL, \citet{hoellmer_open_2026} replace a velocity-based ODE with stochastic surrogate dynamics obtained by adding isotropic Gaussian noise to the numerical integration increment. For sufficiently small noise, this leaves evaluation metrics unchanged. In our case, we inject isotropic Gaussian noise with noise schedule $\sigma_t$ in the tangent space $T_{x_t}\mathcal{M}$ before mapping back to the manifold via the exponential map:
\begin{align}
    x_{t+\Delta t}
    =
    \exp_{x_t}\left(
        \Delta t\, b_t^\theta(x_t)
        +
        \sigma_t \sqrt{\Delta t}\,\xi_t
    \right),
    \qquad
    T_{x_t}\mathcal{M}\ni \xi_t \sim \mathcal{N}(0, I_{\dim\mathcal{M}}).
    \label{eq:wrapped_update}
\end{align}
The resulting conditional probability distribution $\pi^\theta(x_{t+\Delta t}\mid x_t)$ is a wrapped Gaussian on $\mathcal{M}$ \cite{de_surrel_wrapped_2025}, whose log-likelihood factors into a Euclidean Gaussian in the tangent space plus a $\theta$-independent Jacobian term (see Appendix~\ref{sec:rl_appendix} for the factorization; the nontrivial $SO(3)$ and $\mathrm{Sym}_3^+$ Jacobians are given in Appendix~\ref{sec:explogdist}). 

\paragraph{GRPO} The iterative stochastic process in Equation~\ref{eq:wrapped_update} can be understood as a Markov decision process to enable policy-gradient RL~\cite{black2024training}, which aims to optimize the policy $\pi^\theta(x_{t+\Delta t}\mid x_t)$ so that the expected terminal-only reward $r(x_1)$ is maximized. As the reward in our setting, we use the negative all-atom energy from an MLIP---UMA or, in our ablation, Orb---so that generated structures are biased towards smaller energies~\cite{wood_uma_2026,neumann_orb_2024}. GRPO samples $G$ trajectories $\tau^{1:G}$ under identical conditioning---in our case, for the same molecular crystal---and maximizes the following surrogate objective~\cite{shao_deepseekmath_2024}:
\begin{align}
    \mathcal{L}_{\mathrm{GRPO}}(\theta)
    = \tfrac{1}{SGK}\,
    \mathbb{E}_{\tau^{1:G}\sim\pi^{\theta_\mathrm{old}}}\!\left[
        \sum_{i=1}^G \sum_{k=0}^{K-1}
        \min\!\bigl(
            \rho_{i,k}(\theta)\hat A_i,\,
            \mathrm{clip}(\rho_{i,k}(\theta), 1\!-\!\varepsilon, 1\!+\!\varepsilon)\hat A_i
        \bigr)
    \right].
    \label{eq:grpo_loss}
\end{align}
Here, $K$ is the number of integration time steps, $S$ is an optional normalization factor that accounts for the variable atom count across different molecular crystals~\cite{hoellmer_open_2026}, and $\varepsilon$ is a clipping hyperparameter. We further used the group-normalized advantages $\hat A_i = (r(x_1^i) - \operatorname{mean}\{r(x_1^j)\}) / \operatorname{std}\{r(x_1^j)\}$, and the one-step ratio $\rho_{i,k}(\theta) = \pi^\theta(x_{t_{k+1}}^i\mid x_{t_k}^i) / \pi^{\theta_{\mathrm{old}}}(x_{t_{k+1}}^i\mid x_{t_k}^i)$ between the updated policy $\pi^\theta$ and the old policy $\pi^{\theta_{\mathrm{old}}}$ that generated the trajectories. This one-step ratio reduces to a ratio of Euclidean Gaussians in $T_{x_{t_k}}\mathcal{M} \cong \mathbb{R}^{6M+9}$ because, for policies compared at the same base point $x_{t_k}$, the $\theta$-independent Jacobian terms cancel. Besides the objective in Equation~\ref{eq:grpo_loss}, we use a KL-regularization with respect to the pretrained policy $\pi^{\theta_\text{ref}}$, evaluated in closed form between the corresponding tangent-space transitions (Appendix~\ref{sec:rl_appendix}).

\section{Results}

\begin{wraptable}{r}{0.60\textwidth}
  \vspace{-0.8em}
  \centering
  \caption{CCDC packing-similarity metrics on the first 128 structures of the OMC25-MCF test set ($k=30$ inference). Values are rates; $\uparrow$~higher is better and $\downarrow$~lower is better.}
  \label{tab:omc128_results}
  \scriptsize
  \setlength{\tabcolsep}{1.8pt}
  \renewcommand{\arraystretch}{1.15}
  \resizebox{\linewidth}{!}{%
  \begin{tabular}{@{}lcccc@{}}
    \toprule
    Metric & MCF$^\dagger$ & Base & Orb-IRL & UMA-IRL \\
    \midrule
    Solved $\uparrow$ & $0.0391$ & $0.0742{\scriptscriptstyle\,\pm\,0.0039}$ & $0.1008{\scriptscriptstyle\,\pm\,0.0046}$ & $\boldsymbol{0.1273{\scriptscriptstyle\,\pm\,0.0081}}$ \\
    \addlinespace[1.2pt]
    Solved (coll. allowed) $\uparrow$ & $0.1797$ & $0.2109{\scriptscriptstyle\,\pm\,0.0064}$ & $0.2227{\scriptscriptstyle\,\pm\,0.0055}$ & $\boldsymbol{0.2594{\scriptscriptstyle\,\pm\,0.0064}}$ \\
    \addlinespace[1.2pt]
    Packing match $\uparrow$ & $0.4062$ & $0.4680{\scriptscriptstyle\,\pm\,0.0115}$ & $0.5031{\scriptscriptstyle\,\pm\,0.0085}$ & $\boldsymbol{0.5766{\scriptscriptstyle\,\pm\,0.0079}}$ \\
    \addlinespace[1.2pt]
    Packing match/draw $\uparrow$ & $0.0534$ & $0.0679{\scriptscriptstyle\,\pm\,0.0013}$ & $0.0749{\scriptscriptstyle\,\pm\,0.0009}$ & $\boldsymbol{0.0865{\scriptscriptstyle\,\pm\,0.0009}}$ \\
    \addlinespace[1.2pt]
    Clash $\downarrow$ & $0.6182$ & $0.3896{\scriptscriptstyle\,\pm\,0.0022}$ & $0.2716{\scriptscriptstyle\,\pm\,0.0016}$ & $\boldsymbol{0.1732{\scriptscriptstyle\,\pm\,0.0015}}$ \\
    \bottomrule
  \end{tabular}%
  }
  \vspace{0.4em}

  \parbox{0.98\linewidth}{\scriptsize\raggedright For CG-OMatG variants, values are the mean $\pm$ SEM over ten blocks of 30 draws per target (300 draws total). $^\dagger$MCF has one $K=30$ block and therefore no error bars.\par}
  \vspace{-0.6em}
\end{wraptable}

\paragraph{Datasets} \label{sec:data_mainpaper}

We train our model on two separate molecular crystal datasets: \textbf{OMC25-MCF} \citep{gharakhanyan_open_2025}, a subset of the Open Molecular Crystals dataset curated by \citet{zeng_molcrystalflow_2026} containing $46,120$ structures; and \textbf{CSD} \citep{groom_cambridge_2016}, a proprietary dataset of $400,057$ experimentally validated molecular crystal structures maintained by the CCDC and available through the purchase of a license. In both cases, we restrict ourselves to homomolecular crystals and leave cocrystalline and solvated materials as an avenue for future investigation. A complete summary of how the data is preprocessed is provided in Appendix~\ref{app:data_processing}.

\paragraph{Open Molecular Crystals} 

On the first 128 structures of OMC25-MCF---the lowest-\texttt{uma-s-1p1}-energy subset of Open Molecular Crystals \cite{gharakhanyan_open_2025,wood_uma_2026} introduced by \citet{zeng_molcrystalflow_2026}---we compare MCF with the base model and models reinforced using UMA or Orb rewards. The stronger, longer-trained UMA run supplies the main CG-OMatG-IRL results; the shorter Orb run and reward-circularity analysis are discussed in Appendix~\ref{app:additional_results}. The Orb gains show that the improvement is not specific to UMA.

Results are presented in Table~\ref{tab:omc128_results}. To quantify substantial violations of physical interactions, we report clash rates, where a clash occurs if the distance between two intermolecular heavy atoms is less than $0.75$ times the sum of their covalent radii \cite{subramanian_packflow_2026}. We define packing similarity using COMPACK, where a packing match is recorded when at least eight of fifteen molecules in the packing shell can be aligned \cite{motherwell_compack_2005}. Finally, a target is considered solved when the structures are packing similar, possess $\mathrm{RMSD}_{N_{\mathrm{Matches}}}\leq 2.0$\,\AA, and contain no collisions. Collisions are stricter than clashes and occur when the distance between two intermolecular atoms falls below the sum of their van der Waals radii minus $0.7$\,\AA\ \cite{jin_oxtal_2025}. ``Collisions allowed'' applies the same packing and RMSD criteria without the collision screen. Packing match (per draw) and clash are the respective fractions of the 30 draws satisfying the packing-match and clash criteria.
The base model nearly doubles MCF's solved rate, while UMA-RL exceeds three times it and further improves the collision-screened solve rate relative to the base model. Rewards rise and clashes fall during RL on both datasets (Appendix Figure~\ref{fig:rl_reward_curves}), empirically validating the manifold policy-gradient construction and supporting the idea that it distills MLIP physicality into the generator.

\begin{figure}[H]
    \centering
    \includegraphics[width=1\linewidth]{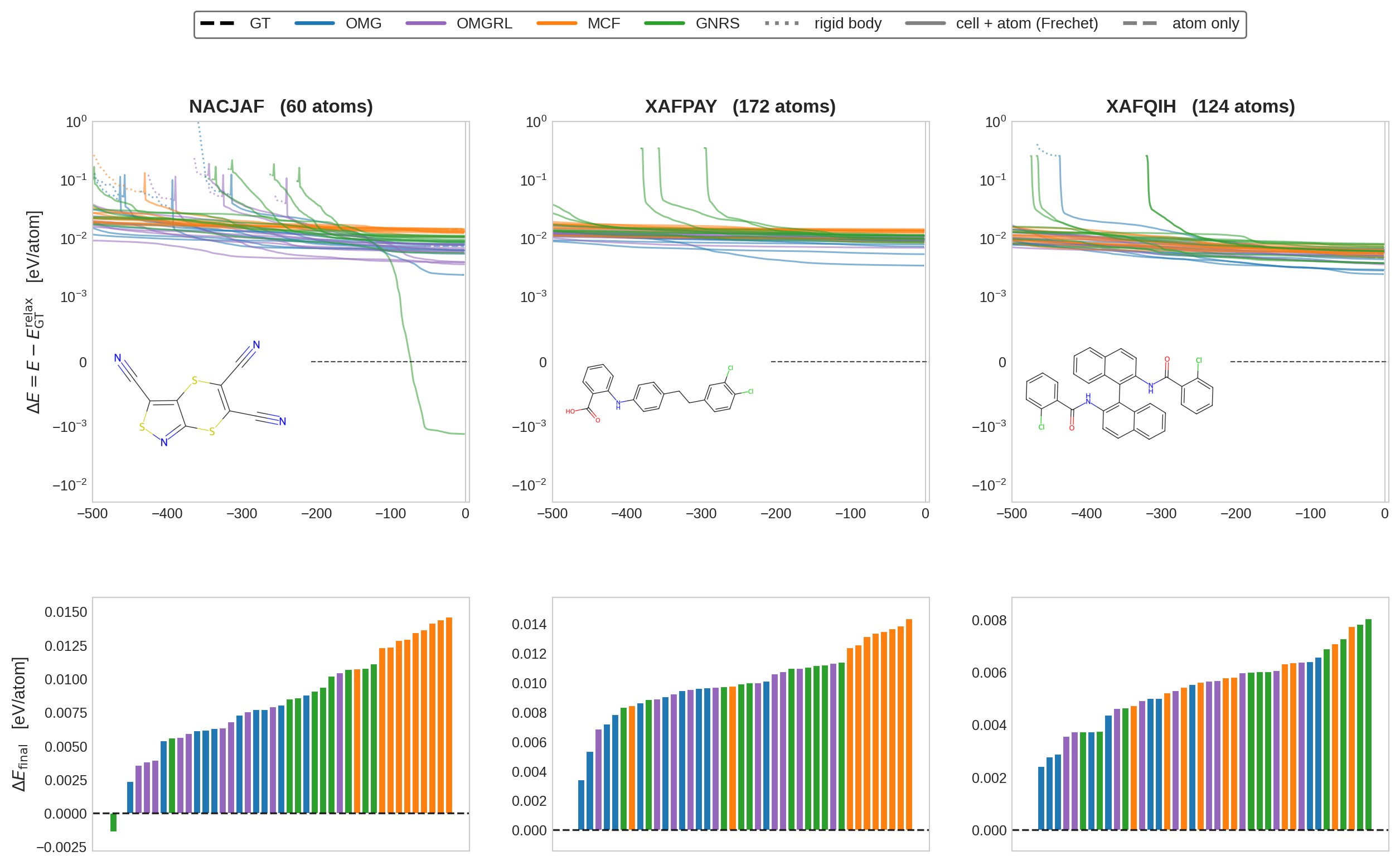}
    \caption{UMA-driven relaxation for the three homomolecular CSP blind-test 6 targets (NACJAF, XAFPAY, XAFQIH), showing the ten lowest-energy draws per generator. Top: energy deviation from the UMA-relaxed ground truth across the three-stage BFGS relaxation; bottom: final deviations sorted by energy. Energies are in eV/atom. Relaxation details and complementary energy--density analyses are in Appendix~\ref{app:relax} and Appendix~\ref{app:additional_results}.}
    \label{fig:energy}
\end{figure}

\paragraph{Cambridge Structural Database}

We benchmark CG-OMatG on CSP blind-test data \cite{groom_cambridge_2016}. The CSP blind test refers to an annual competition hosted by the CCDC in which scientists aim to predict experimentally validated, yet previously unseen, crystal structures. We train CG-OMatG on a large dataset curated from the CSD (Appendix Section~\ref{app:data_processing}). In Figure~\ref{fig:energy}, we demonstrate the performance of CG-OMatG on three crystal targets from the sixth CCDC blind test. We benchmark CG-OMatG against MCF and Genarris 3.0, a popular statistical algorithm for proposing molecular crystal structures \cite{yang_genarris_2025}. Quantitatively, CG-OMatG matches NACJAF before relaxation (8/15 molecules at $1.86$\,\AA) and after relaxation (11/15 at $0.37$\,\AA), and XAFPAY after relaxation (8/15 at $1.51$\,\AA). No method solves XAFQIH, although relaxation improves the best CG-OMatG-IRL match from $2.35$ to $2.03$\,\AA. Appendix~\ref{app:additional_results} reports full blind-test metrics and further studies of velocity annealing, conformer choice, polymorph diversity, and additional benchmark comparisons. In the conformer ablation, ETKDG/MMFF94s sampling recovers conformers within $1$\,\AA\ of the experimental structures, although inference from the generated conformers does not yield an additional solved target.

\section{Discussion}

\paragraph{Methodological Contributions}
To our knowledge, this work is the first to formulate policy-gradient optimization of flow models on a non-trivial Riemannian manifold. This construction circumvents the need to evaluate probability densities using the computationally expensive divergence. 
We additionally introduce a data augmentation scheme that accounts for the symmetries of rigid bodies; because these symmetries arise generically in pose prediction problems, we suspect the approach has implications beyond molecular CSP. Lastly, this work provides the first formal construction of the Riemannian manifold of all possible molecular crystal configurations. These contributions are primarily theoretical and provide a rigorous geometric foundation on which future generative models for molecular crystals and, more generally, periodic rigid-body systems can build. 

\paragraph{Limitations} A current limitation of our approach is that the number of molecules in the unit cell, $M$, must be specified at inference time. In a true blind-test setting where only the molecular graph is given, this requires running the full generation-plus-relaxation pipeline for each candidate number $M$. Additionally, we assume that conformer degrees of freedom factorize and are delta distributed from the crystal packing problem in equation \ref{eq:factorizeprior}. This assumption can be insidious for highly flexible molecules and warrants reexamination in future work. Rigidity applies only during proposal generation: subsequent unconstrained atomistic relaxation can correct moderate conformational errors, but cannot replace explicit conformational sampling. Another limitation of this work is the observed high clash rates. The rigid body approximation, which we exploit in this work, makes molecular crystal systems highly sensitive to improperly learned rotations, which manifest as clashes. These effects are dramatic in systems featuring long, rod-like molecules. Finally, we restrict our study to homomolecular crystals, whereas cocrystals containing multiple distinct molecular species are common in nature and represent an important extension for future work.

\begin{samepage}
\section*{Acknowledgments}
The authors thank the NYU IT High Performance Computing team for their provision of computational resources and general support. The authors acknowledge funding from NSF Grant OAC-2311632. S.\ M.\ acknowledges support from the Simons Center for Computational Physical Chemistry (Simons Foundation grant 839534, MT).  The authors gratefully acknowledge use of the research computing resources of the Empire AI Consortium, Inc., with support from the State of New York, the Simons Foundation, and the Secunda Family Foundation. We thank Shenglong Wang for reserving compute nodes and helping resolve a CCDC license-validation issue.

This material is based upon work supported by the National Science Foundation under Grant Number 2345719. Any opinions, findings, and conclusions or recommendations expressed in this material are those of the author(s) and do not necessarily reflect the views of the National Science Foundation.

Large language models assisted with manuscript drafting and editing and research code development; the authors verified all results and claims and take full responsibility for the work.
\end{samepage}

\bibliography{references}

\newpage

\appendix
\raggedbottom
\startcontents[appendix]
\printcontents[appendix]{}{0}{\section*{Appendix Contents}}

\newpage

\section{The Coarse Graining Map}

\paragraph{Coarse Graining}
Letting an index set $S_i$ specify a subset of atoms constituting a molecule, we reduce the naive degrees of freedom from $\mathbb{R}^{3\times N_i}$ to a coarse-grained descriptor. On each disjoint subset of atoms, the `coarse graining' (CG) map is taken to act equivariantly, producing a pair consisting of the molecular centroid $q^{(i)}\in\mathbb{R}^3$ and orientation $Q^{(i)}\in SO(3)$. 
Following Kim et al.\ \cite{kim_mofflow_2025}, we define the CG mapping in the following way
\begin{definition}{Coarse Graining Map}{cgmap}
    The coarse graining map $\mathcal{C}$ is defined as
    \begin{align*}
        \mathcal{C}: \mathbb{R}^{3\times N_i} &\to \mathbb{R}^3 \times SO(3), \\
        \{c^{(j)}\}_{j\in S_i} &\mapsto (q^{(i)},Q^{(i)}),
    \end{align*}
    where
    \begin{align*}
        q^{(i)} = \frac{1}{N_i}\sum_{j\in S_i} c^{(j)}, \qquad
        Q^{(i)} = \mathcal{R}\left(\{c^{(j)}\}_{j\in S_i}\right),
    \end{align*}
    such that, writing $C=[c^{(1)}|\cdots|c^{(N_i)}]\in\mathbb{R}^{3\times N_i}$ and
    \begin{align*}
        [e_1(C)\mid e_2(C)\mid e_3(C)]
        := \operatorname{Eig}\left(
        \frac{1}{N_i}\sum_{n=1}^{N_i}
        \big(c^{(n)}-q^{(i)}\big)\big(c^{(n)}-q^{(i)}\big)^{\top}
        \right),
    \end{align*}
    where $\operatorname{Eig}(\cdot)$ returns the orthonormal eigenbasis ordered by decreasing eigenvalues, the orientation map is given by
    \begin{align*}
        \mathcal{R}(C)
        :=
        \left[\tilde e_1(C)\ \middle|\ \tilde e_2(C)\ \middle|\ \tilde e_1(C)\times \tilde e_2(C)\right],
        \qquad
        \tilde e_k(C):=\operatorname{sgn}\big(v(C)^\top e_k(C)\big)\,e_k(C),
    \end{align*}
    where $v:\mathbb{R}^{3\times N_i}\to\mathbb{R}^3$ is an auxiliary $SO(3)$-equivariant vector function. Note that this definition is valid only for point clouds in $\mathbb{R}^{3\times N_i}$ that do not admit a stabilizer under any type-preserving orientation, i.e., a non-trivial point group. See Section~\ref{app:symmetries} for our resolution. 
\end{definition}
The mapping $\mathcal{R}$ is (nearly) the principal axis of the molecule; Appendix~\ref{app:symmetries} discusses the associated frame ambiguities.
The `canonical' or `local' coordinates $\tilde{c}^{(j)}$ of atoms in molecule $i$ are then obviously given by
\begin{align}
    \tilde{c}^{(j)} = (Q^{(i)})^\top (c^{(j)} - q^{(i)})
\end{align}
which shifts the molecule to the origin (so its centroid is identically zero) and undoes the rotation $Q^{(i)}$ so the principal components are ordered along the conventional $x,y,z$ axes.

\section{Mathematical Specification of a Molecular Crystal}

\paragraph{Crystals and CSP} We first define a crystal structure mathematically.
This is a key step as we will use this as a foundation to rigorously define a molecular crystal.
\begin{definition}{Crystal}{crystal}
    A crystal $y$ is a tuple
    \begin{align*}
        y:=\left(L,\{c^{(j)}\}_{j=1}^N,\{a^{(j)}\}_{j=1}^N\right),
    \end{align*}
    where $N$ is the total number of atoms in the unit cell, $L\in GL^+(3,\mathbb{R})$ is a row-major matrix specifying the lattice vectors, $c^{(j)}\in\mathbb{R}^3$ is the Cartesian position of atom $j$, and $a^{(j)}\in\{0,1\}^T$ is a one-hot vector over $T$ possible atomic types (e.g., carbon, hydrogen, chlorine). For brevity, we write
    \begin{align*}
        \mathcal{A}:=\{a^{(j)}\}_{j=1}^N.
    \end{align*}
\end{definition}

\paragraph{Molecular Crystals} For molecular crystals we partition the atoms into $M$ disjoint subsets, each constituting a ``molecular conformer'' or ``molecule'' with $N_i$ atoms.
Letting $S_i$ denote the set of atoms in molecule $i$, we have the restrictions
\begin{align}
    \{c^{(j)}\}_{j=1}^N = \bigsqcup_{i=1}^M \{c^{(j)}\}_{j\in S_i},\qquad  \{a^{(j)}\}_{j=1}^N = \bigsqcup_{i=1}^M \{a^{(j)}\}_{j\in S_i},\qquad \sum_{i=1}^M N_i = N,
\end{align}
where $\bigsqcup$ is the disjoint union. 
The first two conditions ensure that the subsets reconstruct the entire crystal; the last condition ensures that all atoms appear once. 
The subsets must be disjoint so that no atom belongs to two conformers. 
It is then natural to define an $M$-molecule coarse-grained crystal as the image of an $N$-atom crystal $y$ under the coarse-graining map, where each set $S_i\in \{S_i\}_{i=1}^M$ is taken to be a connected component of the geometric connectivity graph
\begin{align}
    \mathcal{G}:=(\mathcal{V},\mathcal{E})
    =\big(\{c^{(j)}\}_{j=1}^N,\mathcal{E}\big),
\end{align}
with $\mathcal{V}$ as the set of Cartesian atomic positions defining the nodes and $\mathcal{E}$ as the set of edges encoding chemical bonds. The set of connected components may be denoted $\mathfrak{S}_m:=\{S_i^{(m)}\}_{i=1}^M$. 
\begin{definition}{Coarse Grained Crystal}{cgcrystal}
    A coarse-grained crystal $x$ is a tuple
    \begin{align*}
        x := \big(L,\{q^{(i)},Q^{(i)}\}_{i=1}^M\big),
    \end{align*}
    where $M$ is the number of molecules in the unit cell, $L\in GL^+(3,\mathbb{R})$ is a row-major matrix whose rows are the lattice vectors, $q^{(i)}\in\mathbb{R}^3$ is the Cartesian position of molecule $i$, and $Q^{(i)}\in SO(3)$ is its orientation.

    \ 

    The number of degrees of freedom in a coarse-grained crystal is typically smaller than that of the corresponding atomistic crystal. To retain invertibility of the coarse graining map, it is useful to augment this representation with the local coordinates and atomic species. Accordingly, define
    \begin{align*}
        \tilde{x}
        :=
        \big(L,\{q^{(i)},Q^{(i)}\}_{i=1}^M,C,\mathcal{A}\big),
    \end{align*}
    where
    \begin{align*}
        C
        :=
        \big(\{\tilde c^{(j)}\}_{j\in S_1},\dots,\{\tilde c^{(j)}\}_{j\in S_M}\big),
    \end{align*}
    and where $\mathcal{A}$ denotes the corresponding atomic species.
\end{definition}

\section{Mathematical Definition of a Molecular Crystal Manifold} \label{app:mol_cryst_manifold}
\begin{figure}[H]
    \centering
    \includegraphics[width=1\linewidth]{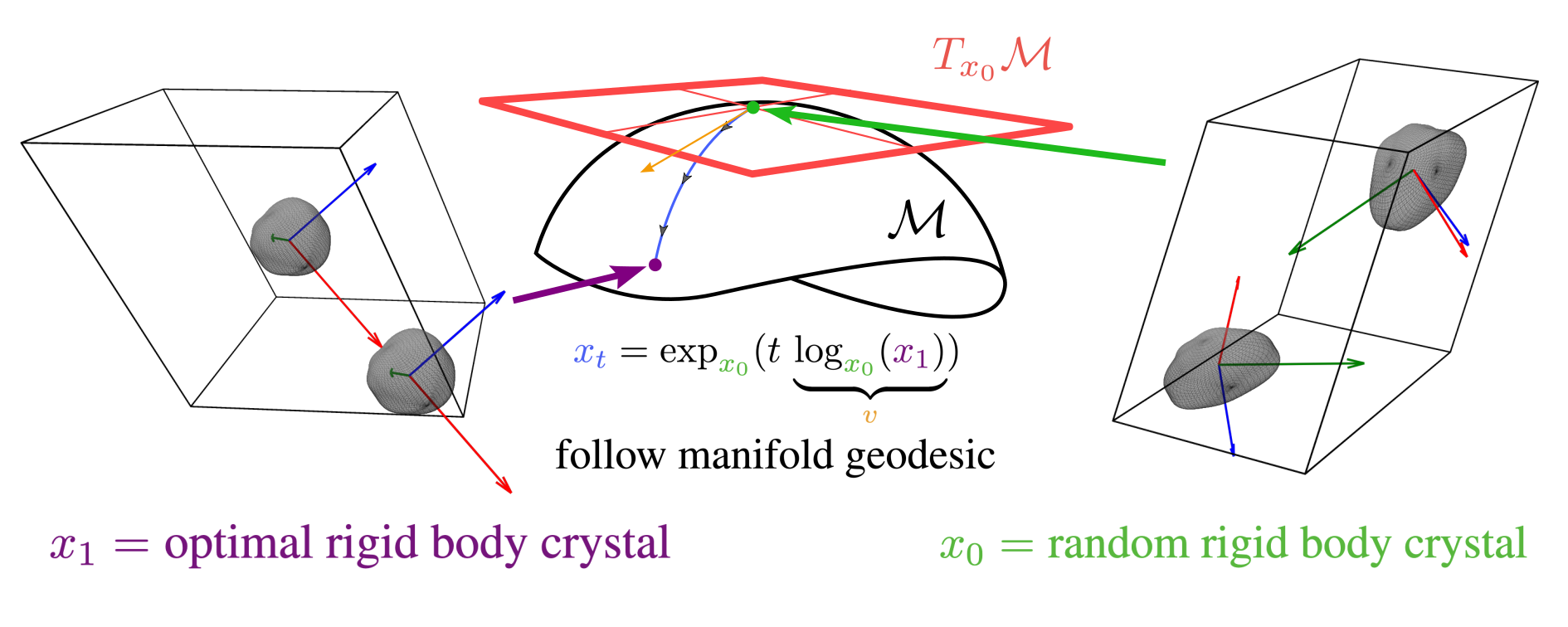}
     \caption{A cartoon of the molecular crystal manifold $\mathcal{M}$. The blue curve represents a geodesic $\textcolor{xtcol}{x_t}
        =
        \exp_{\textcolor{x0col}{x_0}}(
      t
        \log_{\textcolor{x0col}{x_0}}\left(\textcolor{x1col}{x_1}\right)
    )$ connecting an initially random rigid body positions  ${\textcolor{x0col}{x_0}}$ to an optimal set of positions $\textcolor{x1col}{x_1}$. The initial velocity ${\textcolor{vcol}{{v}}}$ is an element of the tangent space $\textcolor{Tx0Mcol}{T_{x_0}\mathcal{M}}$ at the point ${\textcolor{x0col}{x_0}}$. The abstract blobs are rigid bodies with PCA frames attached to them which represent molecules being crystallized.}
    \label{fig:manifold}
\end{figure}

\begin{definition}{Molecular Crystal Manifold}{mcmanifold}
    The molecular crystal manifold $\mathcal{M}$ is the $(6M+9)$-dimensional product manifold
    \begin{align*}
        \mathcal{M} := SO(3)\times \mathrm{Sym}_3^+ \times \left(\mathbb{T}^3 \times SO(3)\right)^M,
    \end{align*}
    endowed with the Riemannian metric $g^{\mathcal{M}}_{x} : T_x\mathcal{M}\times T_x\mathcal{M}  \to \mathbb{R}$ which is additive
    \begin{align*}
        g_x(v_1,v_2)
        &:=
        g_U^{SO(3)}(\mathcal{U}_1,\mathcal{U}_2)
        +
        g_P^{\mathrm{Sym}_3^+}(\mathcal{P}_1,\mathcal{P}_2)
        +
        \sum_{i=1}^M
        \left(
            g_{f^{(i)}}^{\mathbb{T}^3}(\mathcal{F}_1^{(i)},\mathcal{F}_2^{(i)})
            +
            g_{Q^{(i)}}^{SO(3)}(\mathcal{Q}_1^{(i)},\mathcal{Q}_2^{(i)})
        \right),
        \end{align*}
        where the metric on each sub-manifold is given by
        \begin{align*}
        g_U^{SO(3)}(\mathcal{U}_1,\mathcal{U}_2)
        &:= \tfrac{1}{2}\Tr(\mathcal{U}_1^\top \mathcal{U}_2), \qquad
        g_P^{\mathrm{Sym}_3^+}(\mathcal{P}_1,\mathcal{P}_2)
        := \Tr(P^{-1/2}\mathcal{P}_1P^{-1}\mathcal{P}_2P^{-1/2}), \\
        g_f^{\mathbb{T}^3}(\mathcal{F}_1,\mathcal{F}_2)
        &:= \mathcal{F}_1^\top \mathcal{F}_2, \qquad
        g_Q^{SO(3)}(\mathcal{Q}_1,\mathcal{Q}_2)
        := \tfrac{1}{2}\Tr(\mathcal{Q}_1^\top \mathcal{Q}_2).
    \end{align*}
    Here $v_1,v_2\in T_x\mathcal{M}$ are arbitrary tangent vectors at $x$, written as
    \begin{align*}
        v_k = \big(\mathcal{U}_k,\mathcal{P}_k,\mathcal{F}_k^{(1)},\dots,\mathcal{F}_k^{(M)},\mathcal{Q}_k^{(1)},\dots,\mathcal{Q}_k^{(M)}\big),
    \end{align*}
    where
    \begin{align*}
        \mathcal{U}_k \in T_U SO(3), \qquad
        \mathcal{P}_k \in T_P \mathrm{Sym}_3^+, \qquad
        \mathcal{Q}_k^{(i)} \in T_{Q^{(i)}}SO(3), \qquad
        \mathcal{F}_k^{(i)} \in T_{f^{(i)}}\mathbb{T}^3.
    \end{align*}
\end{definition}
The metric on $SO(3)$ is bi-invariant; the metric on $\mathrm{Sym}_3^+$ is affine-invariant. Both admit closed-form geodesics --- Rodrigues' formula \cite{rodrigues_lois_nodate} on $SO(3)$ and Pennec's formula \cite{pennec_riemannian_nodate} on $\mathrm{Sym}_3^+$ --- and the metric on $\mathbb{T}^3$ follows FlowMM \cite{miller_flowmm_2024}. A cartoon of the molecular crystal manifold is shown in figure \ref{fig:manifold}.

\section{Distances, Logarithms, and Exponentials on the Molecular Crystal Manifold} \label{sec:explogdist}

\paragraph{Distances} The distance $d_{\mathcal{M}} : \mathcal{M}\times\mathcal{M} \to \mathbb{R}_{\ge 0}$ between molecular crystal configurations is additive in its square over the product manifold like
\begin{align}
    d^\mathcal{M}(x_1,x_2)^2 = d^{\mathrm{Sym}_3^+}(P_1,P_2)^2 + d^{SO(3)}(U_1,U_2)^2 + \sum_{i=1}^M\left(d^{\mathbb{T}^3}(f_1^{(i)},f_2^{(i)})^2 + d^{SO(3)}(Q_1^{(i)},Q_2^{(i)})^2\right)
\end{align}
where each distance function is given by
\begin{align}
    d^{SO(3)}(Q_1,Q_2) & = \arccos\left(\tfrac{1}{2}\left(\operatorname{Tr}(Q_1^\top Q_2)-1\right)\right), \qquad
    d^{\mathbb{T}^3}(f_1,f_2) = \min_{k\in\mathbb{Z}^3}\|f_1-f_2+k\|, \\
    & d^{\mathrm{Sym}_3^+}(P_1,P_2) = \operatorname{Tr}\!\left(\log\bigl(P_1^{-1/2}P_2P_1^{-1/2}\bigr)\,\log\bigl(P_1^{-1/2}P_2P_1^{-1/2}\bigr)\right)^{1/2}.
\end{align}
These correspond respectively to the minimum-image fractional distance, the relative angle, and the size of the multiplicative deformation taking one positive-definite form to another. Here $\log$ denotes the standard matrix logarithm, defined by spectral decomposition $ \log P = \Lambda\operatorname{diag}(\log\lambda_1,\dots,\log\lambda_n)\Lambda^\top$. 

\paragraph{Logarithms} The Riemannian logarithm $\log_x^{\mathcal{M}} : \mathcal{O} \to T_x\mathcal{M}$ is a componentwise map from an open set $\mathcal{O}$ containing base and target points $x_1$ and $x_2$ to the tangent space at the base point. It is given by
\begin{align}
    \log_{x_1}^{\mathcal{M}}( x_2) = \left(\log_{U_1}^{SO(3)}(U_2),\log_{P_1}^{\mathrm{Sym}_3^+}(P_2),\{\log_{f^{(i)}_1}^{\mathbb{T}^3}(f_2^{(i)}),\log_{Q_1^{(i)}}^{SO(3)}(Q_2^{(i)})\}_{i=1}^M\right).
\end{align}
where 
\begin{align}
    \log_{Q_1}^{SO(3)}(Q_2) &= Q_1\left(\frac{d^{SO(3)}(Q_1,Q_2)}{2\sin d^{SO(3)}(Q_1,Q_2)}\left(Q_1^\top Q_2-Q_2^\top Q_1\right)\right), \qquad
    \log_{f_1}^{\mathbb{T}^3}(f_2) = f_2-f_1-n^\star, \\
    & \qquad \qquad \qquad \log_{P_1}^{\mathrm{Sym}_3^+}(P_2) = P_1^{\frac12}\log\bigl(P_1^{-\frac12}P_2P_1^{-\frac12}\bigr)P_1^{\frac12},
\end{align}
The value $n^\star \in \arg\min_{n\in\mathbb{Z}^3}\|(f_2-f_1)-n\|$, which implies logarithm on the torus always points from $f_1$ to the nearest periodic image of $f_2$, meaning it can be multivalued when they are half a period apart. Similarly, the logarithm on $SO(3)$ is multivalued near relative angles $\pi$.

\paragraph{Exponentials} The Riemannian exponential $  \exp^{\mathcal{M}}_x : T_x\mathcal{M} \longrightarrow \mathcal{M}$ gives the result of following the geodesic defined by the tangent vector for one unit of time. This is given by
\begin{align}
    \exp_{x}^{\mathcal{M}}(v) = \left(\exp_{U}^{SO(3)}(\mathcal{U}),\exp_{P}^{\mathrm{Sym}_3^+}(\mathcal{P}),\{\exp_{f^{(i)}}^{\mathbb{T}^3}(\mathcal{F}^{(i)}),\exp_{Q^{(i)}}^{SO(3)}(\mathcal{Q}^{(i)})\}_{i=1}^M\right),
\end{align}
where on each submanifold we have
\begin{align}
    \exp_{Q}^{SO(3)}(\mathcal{Q}) & = Q\left(I + \frac{\sin \theta}{\theta}(Q^\top\mathcal{Q}) + 2\frac{\sin^2(\theta/2)}{\theta^2}(Q^\top\mathcal{Q})^2\right), \qquad
    \theta = \sqrt{-\tfrac{1}{2}\operatorname{Tr}\!\left((Q^\top\mathcal{Q})^2\right)}, \\
    & \exp_{P}^{\mathrm{Sym}_3^+}(\mathcal{P}) = P^{\frac12}\exp\!\bigl(P^{-\frac12}\mathcal{P}P^{-\frac12}\bigr)P^{\frac12}, \qquad
    \exp_{f}^{\mathbb{T}^3}(\mathcal{F}) = f+\mathcal{F}\ \mathrm{mod}\ \mathbb{Z}^3.
\end{align}

\paragraph{Jacobians} The Jacobian determinant of $\exp_{P}^{\mathrm{Sym}_3^+}$ is \cite{bhatia_positive_2007,de_surrel_wrapped_2025,chevallier_wrapped_2021}
\begin{align}
    J^{\mathrm{Sym}_3^+}_{P}(\mathcal{P}) = 2^{d(d-1)/2}\prod_{i<j}\frac{\sinh\!\bigl(\tfrac{s_i-s_j}{2}\bigr)}{s_i - s_j}, \qquad s_i = \lambda_i\!\left(P^{-\frac12}\mathcal{P}\,P^{-\frac12}\right).
\end{align}
The Jacobian determinant of $\exp_{Q}^{SO(3)}$ is \cite{chirikjian_stochastic_2012,sola_micro_2021}
\begin{align}
    J^{SO(3)}_{Q}(\mathcal{Q}) = \frac{2(1-\cos\theta)}{\theta^2}, \qquad \theta = \sqrt{-\tfrac{1}{2}\operatorname{Tr}\!\left((Q^\top\mathcal{Q})^2\right)}.
\end{align}
On $\mathbb{T}^3$, $J^{\mathbb{T}^3}_{f}(\mathcal{F}) = 1$ as it is flat.

\section{Insufficiency of a Euclidean Unit Cell} \label{app:euc_unitcell}

\paragraph{Euclidean Description.} We briefly comment on the possibility for a flow prescribed by interpolation on $\mathbb{R}^{3\times3}$ to leave the manifold $GL^+(3,\mathbb{R})$. Consider the two cell matrices
\begin{align}
    L_0 = \begin{bmatrix}
        1 & 0 & 0 \\
        0 & 1 & 0 \\
        0 & 0 & 1
    \end{bmatrix}, \quad L_1 = \begin{bmatrix}
        -1 & 0 & 0 \\
        0 & -2 & 0 \\
        0 & 0 & 3
    \end{bmatrix}.
\end{align}
In either case the determinant is positive: $\det L_0 = 1$, $\det L_1 = 6$, so both are valid endpoints for our interpolation. Now consider their linear interpolation:
\begin{align}
    L_t = (1-t) L_0 + t L_1 \implies L_{t=0.4} = \begin{bmatrix}
        0.2 & 0 & 0 \\
        0 & -0.2 & 0 \\
        0 & 0 & 1.8
    \end{bmatrix}.
\end{align}
This yields $\det L_{t=0.4} = -0.072 < 0$, meaning the interpolant has left $GL^+(3,\mathbb{R})$ and no longer defines a valid unit cell. Naive linear interpolation in $\mathbb{R}^{3\times 3}$ does not respect the topology of the constraint set. One alternative is to canonicalize all cell matrices into a standard orientation before interpolation, e.g.\ by extracting lattice parameters and reconstructing a lower-triangular cell in a fixed frame as in Crystalite \cite{veljkovic_crystalite_2026} or applying a Niggli reduction to all the data during preprocessing.

\paragraph{Our Solution.} We decompose the cell via $L = UP$ into a rotation $U \in SO(3)$ and a symmetric positive-definite stretch $P \in \mathrm{Sym}_3^+$, and interpolate each factor along geodesics of its intrinsic Riemannian metric. Since both $SO(3)$ and $\mathrm{Sym}_3^+$ are geodesically complete, the interpolant remains on the manifold for all $t \in [0,1]$ by construction. Every intermediate point is a valid rotation composed with a valid positive-definite stretch, and hence a valid element of $GL^+(3,\mathbb{R})$.

\section{Reinforcement Learning on The Molecular Crystal Manifold}\label{sec:rl_appendix}

This appendix expands the manifold-RL construction summarized in Section~\ref{sec:rl_main}

\paragraph{Markov Decision Process} Following \citet{hoellmer_open_2026}, we use reinforcement learning (RL) as a post-training fine-tuning step to improve transferability and to bias generation toward energetically stable crystal structures. Our setting extends their construction from Euclidean generative dynamics to the manifold-valued dynamics of molecular crystals. 

First, we cast the time-discretized dynamics on $\mathcal{M}$ as a Markov decision process
\begin{align}
    (\mathcal{X}, \mathcal{Y}, \mu_0, K, R),
\end{align}
with state space $\mathcal{X} = [0,1] \times \mathcal{M}$, action space $\mathcal{Y}$, initial-state distribution $\mu_0 = (\delta_0, p_0)$, transition kernel $K$, and reward function $R$. At time $t$, the state is $s_t = (t, x_t)\in\mathcal{X}$
and the initial state is drawn as $s_0 \sim \mu_0$, so every trajectory begins at $t = 0$ from a prior sample $x_0 \sim p_0$. The agent samples an action from the stochastic policy
\begin{align}
    \pi^\theta(a_t \mid s_t) := \pi^\theta(x_{t+\Delta t} \mid x_t),
\end{align}
which we identify with the next configuration, $a_t := x_{t+\Delta t}$. The transition kernel is deterministic given the action,
\begin{align}
    K(s_{t+\Delta t} \mid s_t, a_t) = \delta_{(t+\Delta t,\, a_t)},
\end{align}
so all stochasticity in the trajectory comes from the policy itself. We take the reward to be terminal-only,
\begin{align}
    R(s_t, a_t)
    :=
    \begin{cases}
        r(x_1) & \text{if } t = 1,\\
        0 & \text{otherwise},
    \end{cases}
\end{align}
and choose
$
    r(x_1) = - E(x_1),
$
where $E$ is the all-atom energy of the final crystal computed using UMA \cite{wood_uma_2026}. This energy calculation includes periodic boundary conditions from the predicted unit cell $L$.

\paragraph{Promotion of the ODE to an SDE} In the base formulation, the learned generative dynamics define a deterministic ODE, so the induced policy is likewise deterministic. This is undesirable for RL, where some degree of stochasticity is needed for exploration. \citet{hoellmer_open_2026} address this in the flat setting by augmenting the dynamics with controlled noise. Here we extend that idea to the curved manifold $\mathcal{M}$ by introducing noise directly in the tangent space $T_{x_t}\mathcal{M}$. 

Given a trained velocity field $b_t^\theta$, deterministic integration is performed by a Riemannian Euler step with step size $\Delta t$ using the exponential map as the local chart:
\begin{align}
    x_{t+\Delta t}
    =
    \exp_{x_t}\left(\Delta t\, b_t^\theta(x_t)\right).
\end{align}
To obtain a stochastic policy, we instead add isotropic Gaussian noise in the tangent space before mapping the update back to the manifold,
\begin{align}
    x_{t+\Delta t}
    =
    \exp_{x_t}\left(
        \Delta t\, b_t^\theta(x_t)
        +
        \sigma_t \sqrt{\Delta t}\,\xi_t
    \right),
    \qquad
    \xi_t \sim \mathcal{N}(0, I_{\dim\mathcal{M}}).
    \label{eq:wrapped_update_app}
\end{align}
Equivalently, if we denote the tangent-space increment by
$w_t := \Delta t\, b_t^\theta(x_t) + \sigma_t \sqrt{\Delta t}\, \xi_t \in T_{x_t}\mathcal{M}$,
then $w_t$ is Gaussian in $T_{x_t}\mathcal{M}$ with mean $\Delta t\, b_t^\theta(x_t)$ and covariance $\sigma_t^2 \Delta t\, I_{\dim\mathcal{M}}$, and the next state is obtained by the pushforward $x_{t+\Delta t} = \exp_{x_t}^{\mathcal{M}}(w_t)$. The resulting policy is therefore a wrapped Gaussian on $\mathcal{M}$ \cite{de_surrel_wrapped_2025}. 

Using the inverse map $w_t = \log_{x_t}^{\mathcal{M}}(x_{t+\Delta t})$, the usual change-of-variables formula first gives
\begin{align}
    \pi^\theta(x_{t+\Delta t} \mid x_t)
    &=
    \mathcal{N}\left(
        \log_{x_t}^{\mathcal{M}}(x_{t+\Delta t})
        \,\middle|\,
        \Delta t\, b_t^\theta(x_t),
        \sigma_t^2 \Delta t\, I_{\dim\mathcal{M}}
    \right)
    \left|
        \det d\log_{x_t}^{\mathcal{M}}(x_{t+\Delta t})
    \right|.
\end{align}
Since $d\log_{x_t}^{\mathcal{M}}$ is the inverse of $d\exp_{x_t}^{\mathcal{M}}$, this may equivalently be written as
\begin{align}
    \pi^\theta(x_{t+\Delta t} \mid x_t)
    =
    \frac{
        \mathcal{N}\left(
            \log_{x_t}^{\mathcal{M}}(x_{t+\Delta t})
            \,\middle|\,
            \Delta t\, b_t^\theta(x_t),
            \sigma_t^2 \Delta t\, I_{\dim\mathcal{M}}
        \right)
    }{
        J_{x_t}^{\mathcal{M}}\left(\log_{x_t}^{\mathcal{M}}(x_{t+\Delta t})\right)
    },
\end{align}
where $J_{x_t}^{\mathcal{M}}(w) = |\det d\exp_{x_t}^{\mathcal{M}}(w)|$ is the Jacobian of the exponential map at base point $x_t$. Taking logarithms yields
\begin{align}
    \log \pi^\theta(x_{t+\Delta t} \mid x_t)
    = & 
    -\frac{\dim\mathcal{M}}{2}\log\left(2\pi\sigma_t^2\Delta t\right)
    -
    \frac{
        \left\|
            \log_{x_t}^{\mathcal{M}}(x_{t+\Delta t}) - \Delta t\, b_t^\theta(x_t)
        \right\|^2
    }{
        2\sigma_t^2\Delta t
    } \nonumber \\
   &  \quad -
    \log J_{x_t}^{\mathcal{M}}\left(\log_{x_t}^{\mathcal{M}}(x_{t+\Delta t})\right).
    \label{eq:wrapped_logp}
\end{align}
Because $\mathcal{M}$ is a product manifold, this Jacobians determinant factorizes over its component manifolds. In particular, the torus contribution is trivially the identity as it is flat, while the nontrivial geometric corrections come from the $SO(3)$ and $\mathrm{Sym}_3^+$ factors. These factors are provided in Appendix~\ref{sec:explogdist}.

\paragraph{Reinforcement Learning Objective} Policy-gradient RL aims to maximize the expected terminal reward under trajectories $\tau$ generated by the policy,
\begin{align}
    \mathcal{J}[\pi^\theta]
    =
    \mathbb{E}_{\tau \sim \pi^\theta}\left[r(x_1)\right].
\end{align}
As discussed in the main text, in practice, GRPO samples $G$ trajectories $\tau^{1:G}$ from the stochastic policy under identical conditioning and maximizes the following clipped surrogate objective~\cite{shao_deepseekmath_2024}:
\begin{align}
    \mathcal{L}_{\mathrm{GRPO}}(\theta)
    =
    \frac{1}{SGK}
    \mathbb{E}_{\tau^{1:G}\sim\pi^{\theta_\mathrm{old}}}\left[
        \sum_{i=1}^G
        \sum_{k=0}^{K-1}
        \min\left(
            \rho_{i,k}(\theta)\hat A_i,
            \mathrm{clip}\left(\rho_{i,k}(\theta), 1-\varepsilon, 1+\varepsilon\right)\hat A_i
        \right)
    \right].
\end{align}
Here, $K$ is the number of integration time steps, $S$ is an optional normalization factor that accounts for different system sizes across different GRPO groups~\cite{hoellmer_open_2026}, and $\varepsilon$ is a clipping hyperparameter. For $G$ sampled trajectories with rewards $\{r_i\}_{i=1}^G=\{r(x^i_1)\}_{i=1}^G$, the group-relative advantages are given by
\begin{align}
    \hat A_i
    =
    \frac{r_i - \operatorname{mean}(\{r_j\}_{j=1}^G)}{\operatorname{std}(\{r_j\}_{j=1}^G)}.
\end{align}
Since the reward is terminal-only, this same normalized score is used across all steps of trajectory $x^i_t$ in the clipped surrogate objective.
The one-step likelihood ratio between the updated policy $\pi^\theta$ and the old policy $\pi^{\theta_\mathrm{old}}$ that generated the trajectories is given by
\begin{align}
    \rho_{i,k}(\theta)
    :=
    \frac{\pi^\theta(x^i_{t_{k+1}} \mid x^i_{t_k})}{\pi^{\theta_{\mathrm{old}}}(x^i_{t_{k+1}} \mid x^i_{t_k})}
    =
    \frac{
        \mathcal{N}\left(
            \log_{x^i_{t_k}}^{\mathcal{M}}(x^i_{t_{k+1}})
            \,\middle|\,
            \Delta t\, b_t^\theta(x^i_{t_k}),
            \sigma_t^2 \Delta t\, I_{\dim\mathcal{M}}
        \right)
    }{
        \mathcal{N}\left(
            \log_{x^i_{t_k}}^{\mathcal{M}}(x^i_{t_{k+1}})
            \,\middle|\,
            \Delta t\, b_t^{\theta_{\mathrm{old}}}(x^i_{t_k}),
            \sigma_t^2 \Delta t\, I_{\dim\mathcal{M}}
        \right)
    }.
\end{align}
Here, the Jacobian factor from the wrapped Gaussian policy cancels exactly, since it depends only on the manifold geometry and not on $\theta$ due to the tangent covariance of the policy being fixed. 

\paragraph{Kullback--Leibler Regularization} To prevent the updated policy from drifting too far from the reference policy $\pi^{\theta_\text{ref}}$ of the pretrained model, we additionally include a KL regularization term penalizing large deviations of the updated drift $b_t^\theta$ from the reference at each step. Writing the KL contribution for a single rollout trajectory, we add the following term to the maximized objective:
\begin{align}
    \mathcal{L}_{\mathrm{KL}}(\theta)
    =-
    \beta\,
    \mathbb{E}_{\tau \sim \pi^{\theta_{\mathrm{old}}}}\left[
        \sum_{k=0}^{K-1}
        D_{\mathrm{KL}}\left(
            \pi^{\theta}(\cdot \mid x_{t_k})
            \,\|\,
            \pi^{\theta_{\mathrm{ref}}}(\cdot \mid x_{t_k})
        \right)
    \right],
\end{align}
where $\beta$ controls the strength of the regularization. For a single step $k$, substituting the wrapped Gaussian form of the policy and writing $w = \log_{x_{t_k}}^{\mathcal{M}}(x)$ gives
\begin{align}
    & D_{\mathrm{KL}}\left(
        \pi^\theta(\cdot \mid x_{t_k})
        \,\|\,
        \pi^{\theta_{\mathrm{ref}}}(\cdot \mid x_{t_k})
    \right)
    \\&= 
    \int_{\mathcal{M}}
    \frac{
        \mathcal{N}(w \mid \Delta t\, b_{t_k}^{\theta}(x_{t_k}), \sigma_{t_k}^2 \Delta t\, I)
    }{
        J_{x_{t_k}}^{\mathcal{M}}(w)
    }
    \log
    \frac{
        \mathcal{N}(w \mid \Delta t\, b_{t_k}^{\theta}(x_{t_k}), \sigma_{t_k}^2 \Delta t\, I)
    }{
        \mathcal{N}(w \mid \Delta t\, b_{t_k}^{\theta_\mathrm{ref}}(x_{t_k}), \sigma_{t_k}^2 \Delta t\, I)
    }
    \, d\mathrm{vol}_{\mathcal{M}}(x)
    \\
    &=
    \int_{T_{x_{t_k}}\mathcal{M}}
    \mathcal{N}(w \mid \Delta t\, b_{t_k}^{\theta}(x_{t_k}), \sigma_{t_k}^2 \Delta t\, I)
    \log
    \frac{
        \mathcal{N}(w \mid \Delta t\, b_{t_k}^{\theta}(x_{t_k}), \sigma_{t_k}^2 \Delta t\, I)
    }{
        \mathcal{N}(w \mid \Delta t\, b_{t_k}^{\theta_\mathrm{ref}}(x_{t_k}), \sigma_{t_k}^2 \Delta t\, I)
    }
    \, dw,
\end{align}
where in the second line we used $d\mathrm{vol}_{\mathcal{M}}(x) = J_{x_{t_k}}^{\mathcal{M}}(w)\, dw$. Thus, for policies compared at the same base point $x_{t_k}$, the manifold Jacobian cancels exactly, and the KL reduces to the ordinary Euclidean KL between the corresponding tangent-space Gaussians.

\section{Symmetries of Molecular Crystals} \label{app:symmetries}

\paragraph{Lattice Translations} The defining feature of a crystal is its translational symmetry, which presents as a discrete translational symmetry \textit{via} the lattice vectors of the crystalline unit cell, or periodic repeating unit. 
Any crystal is invariant under the action of the lattice translation group
\begin{equation}
    \mathcal{T}(L) = \bigl\{n_1l_x
        + n_2l_y + n_3l_z \;\big|\; n\in\mathbb{Z}^3\bigr\}
\end{equation}
Consequently, positions are only physically meaningful modulo the lattice, i.e.\ as fractional coordinates $f^{(i)} = \operatorname{wrap}(c^{(i)} L^{-1}) \in \mathbb{T}^3$. 

In principle, this means the fractional centroid of the lattice point can be represented with any real number---as long as it is understood that points in 3D space are equivalent under lattice translations. To see this, let $\sim$ be a relation on the set $\mathbb{R}^3$ given by
\begin{align}
   \forall x,y\in \mathbb{R}^3:  {x} \sim {y} \iff ({x}  - {y}){L}^{-1} \in \mathbb{Z}^3 \label{eq:equiv}
\end{align}
where ${L} \in GL^+(3,\mathbb{R})$ are the unit-cell lattice vectors.

\textbf{Claim.} The lattice translation $\sim$ is a valid equivalence relation.

\textbf{Proof.} To be a valid equivalence relation it must be reflexive, symmetric, and transitive. 
\begin{itemize}[leftmargin=*,itemsep=1pt,topsep=2pt,parsep=0pt]
    \item \textbf{Reflexivity}: Observe that $\forall {x} \in\mathbb{R}^3 : {x}  - {x} = {0}$. 
    For any matrix the following holds: $ \forall {A} \in \mathbb{R}^{3\times3}:{0}{A} ={0}$. 
    This implies that $({x}  - {x}){L^{-1}} = {0} \in \mathbb{Z}^3$ which in turn implies that $\forall {x}\in \mathbb{R}^3: {x}\sim {x}$, showing the relation is indeed reflexive.
    \item \textbf{Symmetry}: If ${x}\sim {y}$ then $({x}  - {y}){L}^{-1} \in \mathbb{Z}^3$. 
    The negation $-({x}  - {y}){L}^{-1} \in \mathbb{Z}^3$ is true because $\mathbb{Z}^3$ with $+$ operation forms a group, and group elements have inverses. 
    Accordingly $-({x}  - {y}){L}^{-1} = ({y}  - {x}){L}^{-1}\iff {y} \sim {x}$, showing that the relation is indeed symmetric. 
    \item \textbf{Transitivity}: Supposing $({x}  - {y}){L}^{-1} = {m} \in \mathbb{Z}^3$ and $({y}  - {z}){L}^{-1} = {n} \in \mathbb{Z}^3$ implies
    ${y} = {n}{L} + {z}$. Substituting this into the relation ${x}\sim {y}$ gives $({x} -({n}{L} + {z})){L}^{-1} = ({x} -{z}){L}^{-1} - {n} = {m} \implies ({x}-{z}){L}^{-1}=n+m\in\mathbb{Z}^3$, showing that even after substitution this remains in $\mathbb{Z}^3$. Because $\mathbb{Z}^3$ forms a group, ${m} + {n} \in \mathbb{Z}^3$. Therefore ${x}\sim {y} \land {y}\sim {z} \implies {x}\sim {z}$, showing that the relation is indeed transitive.
\end{itemize}
Therefore the relation defined by Equation~\ref{eq:equiv} is an equivalence relation. \hfill $\square$

Define the set
\begin{align}
    [{x}] = \{{y}\in\mathbb{R}^3| {x} \sim {y}\}
\end{align}
as the lattice equivalence class. Similarly, a location in the lattice is defined as an element of the quotient set\footnote{Note that all elements $[{x}]$ are in the power set $\mathcal{P}(\mathbb{R}^3)$; however, not all elements of the power set are valid equivalence classes. The set $[{x}]$ must satisfy ${x} \sim {y}$ for all ${x},{y} \in [{x}]$. The epsilon relation is placed here to ensure we satisfy restricted comprehension.}
\begin{align}
    \mathbb{R}^3/{L}  := \left\{[{x}] \in \mathcal{P}(\mathbb{R}^3)| {x} \in \mathbb{R}^3\right\} 
\end{align}
which emphasizes that ``a location in the lattice'' is a set of points representing the infinite periodic point pattern termed a crystal. This is the same point made before Equation~\ref{eq:equiv}.

In this work we choose the unit cell as the representative. 
The choice is degenerate because many unit cells can map to the same crystal. We express this choice in fractional coordinates via the wrapping function:
\begin{align}
    f^{(i)} = \operatorname{wrap}(c^{(i)} L^{-1}) \in \mathbb{T}^3
\end{align}
where the wrap always gives $f^{(i)} \in [0,1)$. We will call the following map
\begin{align}
    \pi: \mathbb{R}^3 & \to  \mathbb{R}^3/{L}  \\
    {c}^{(j)} & \mapsto \pi(c^{(j)}) = [c^{(j)}]
\end{align}
the quotient map. This map takes a given point ${c^{(j)}}$ to its equivalence class. By choosing a consistent representative in this way we can ensure the generative model always trains on examples from the same type of representative. Mathematically this is similar to the canonicalization choice made for symmetric point clouds in the paper \cite{levy-jurgenson_manifold_2026}.

With respect to this group, if we were to apply its symmetry operation (which in this case is associated with the integer translation $n\in\mathbb{Z}^3$) the resulting transformation $\Phi_g$ on each component of a point $x=\left(U, P, \{f^{(i)}, Q^{(i)}\}_{i=1}^{M}\right)\in \mathcal{M}$ would be
\begin{align}
    \Phi_g: U\to U,\quad  P\to P,\quad  f^{(i)}\to f^{(i)} + n, \quad Q^{(i)}\to Q^{(i)}
\end{align}
Since $\Phi_g$ shifts $f^{(i)}$ by a constant $n\in\mathbb{Z}^3$ and acts as the identity on all other components, the differential acts trivially on the tangent space:
\begin{align}
    d\Phi_g: \dot{U}\to \dot{U},\quad \dot{P}\to \dot{P},\quad \dot{f}^{(i)}\to \dot{f}^{(i)}, \quad \dot{Q}^{(i)}\to \dot{Q}^{(i)}
\end{align}
This means our network must be invariant to integer translations. We achieve this by depending strictly on Cartesian-space differences $c^{(j)} - c^{(j')}$ and fractional-coordinate differences $\operatorname{wrap}(f^{(j)} - f^{(j')})$. 

\paragraph{Translations} Beyond the discrete lattice, a physically correct model must also respect \emph{global} (rigid-body) continuous translations $c^{(j)} \mapsto c^{(j)} + \tau$, $\tau \in \mathbb{R}^3$, which shift the entire crystal without changing interatomic distances.
The group of global translations is $(\mathbb{R}^3, +)$, and it acts trivially on our parameterization of the manifold point. Applying $\Phi_g$ for $g = \tau \in \mathbb{R}^3$:
\begin{align}
    \Phi_g: U\to U,\quad P\to P,\quad f^{(i)}\to f^{(i)} + \tau L^{-1}, \quad Q^{(i)}\to Q^{(i)}
\end{align}
Since $\tau L^{-1}$ is a constant shift in fractional coordinates, the differential again acts trivially:
\begin{align}
    d\Phi_g: \dot{U}\to \dot{U},\quad \dot{P}\to \dot{P},\quad \dot{f}^{(i)}\to \dot{f}^{(i)}, \quad \dot{Q}^{(i)}\to \dot{Q}^{(i)}
\end{align}
Invariance to continuous translations is achieved in the same manner as lattice translations: through a dependence on strictly pairwise differences $c^{(j)} - c^{(j')}$ and $\operatorname{wrap}(f^{(j)} - f^{(j')})$, in which the uniform shift $\tau$ cancels identically.

\paragraph{Rotations}
In addition to translations, physically meaningful properties of a crystal are invariant under global rotations. Because $L$ is row-major, a global rotation $R \in SO(3)$ acts as $c^{(j)} \mapsto c^{(j)}R^\top$, or equivalently $L \mapsto LR^\top$. Since $L^\top = UP$ is the polar decomposition with $U \in SO(3)$ and $P$ symmetric positive-definite, this gives
\begin{align}
    L^\top \mapsto (LR^\top)^\top = RL^\top = RUP
\end{align}
Since $RU \in SO(3)$ and $P$ is unchanged, uniqueness of the polar decomposition implies $U \mapsto RU$ and $P \mapsto P$. The fractional coordinates are also unchanged:
\begin{align}
    f^{(i)} = c^{(i)}L^{-1} \mapsto c^{(i)}R^\top (LR^\top)^{-1} = c^{(i)}R^\top R L^{-1} = c^{(i)}L^{-1} = f^{(i)}
\end{align}
Applying $\Phi_g$ for $g = R \in SO(3)$:
\begin{align}
    \Phi_g: U \to RU, \quad P \to P, \quad f^{(i)} \to f^{(i)}, \quad Q^{(i)} \to RQ^{(i)}
\end{align}
The differential is:
\begin{align}
    d\Phi_g: \dot{U} \to R\dot{U}, \quad \dot{P} \to \dot{P}, \quad \dot{f}^{(i)} \to \dot{f}^{(i)}, \quad \dot{Q}^{(i)} \to R\dot{Q}^{(i)}
\end{align}
The rotation acts nontrivially on the orientation factors $U$ and $Q^{(i)}$, while leaving the invariant components $P$ and $f^{(i)}$ unchanged. Our network is constructed to be explicitly equivariant with respect to these transformation laws. See Section~\ref{app:architecture} for details. 

\paragraph{Molecular Point Group Symmetries}
To give an example before stating the formal proofs, consider water (H$_2$O): its two hydrogens are related by a $180^\circ$ rotation ($C_2$) about the bisector axis. Suppose we extract an orientation frame $Q$ by PCA on the atomic positions. Now rotate the entire molecule by this $C_2$ rotation. The two hydrogens swap, but since they are identical atoms, the resulting configuration is physically indistinguishable from the original. Yet the frame has rotated: because PCA is equivariant \cite{li_closer_2021}, the rotated configuration yields $g\,Q \neq Q$. Both $Q$ and $g\,Q$ are equally valid orientations of the same physical molecule, and there is no principled way to prefer one over the other. Any single-valued map that tries to do so will violate equivariance. The impossibility proof below makes this precise; the resolution is to return \emph{both} orientations (more generally, the full orbit under the molecular point group) rather than choosing one.

Formally, let $c^{(j)} \in \mathbb{R}^3$ for $j \in S_i$ denote the Cartesian positions of the $N_i$ atoms in molecule $i$. Define the \emph{rotational point group} $\mathcal{G}_i \subset SO(3)$ as the set of rotations under which the molecular configuration is physically indistinguishable:
\begin{align}
    \mathcal{G}_i = \{g \in SO(3) \mid g \cdot C \text{ is identical to } C \text{ up to relabeling of identical atoms}\}.
\end{align}
This is precisely the stabilizer of the physical configuration: the subgroup of $SO(3)$ that fixes the molecule as an unordered point cloud of labeled species. For water, $\mathcal{G}_i = \{I,\, C_2\}$.

\textbf{Claim.} If $\mathcal{G}_i$ is nontrivial, there is no single-valued $SO(3)$-equivariant map from the physical configuration to $SO(3)$.

\textbf{Proof.} Suppose such a map $\mathcal{R}$ exists and denote its value $Q^{(i)} = \mathcal{R}(C) \in SO(3)$. Let $g \in \mathcal{G}_i$ with $g \neq I$. Since $g$ fixes the physical configuration, $\mathcal{R}(g \cdot C) = \mathcal{R}(C)$. By equivariance, $\mathcal{R}(g \cdot C) = g\,Q^{(i)}$. Together:
\begin{align}
    g\,Q^{(i)} = Q^{(i)}.
\end{align}
Right-multiplying by $(Q^{(i)})^{-1}$ yields $g = I$, contradicting $g \neq I$. Therefore the map is either not equivariant or not single-valued. \hfill $\square$

The preceding result implies that any molecule admitting a nontrivial permutation stabilizer realized by a rotation cannot have a single-valued equivariant orientation map. The coarse-graining map in Def.~\ref{def:cgmap} must instead be set-valued, returning the \emph{orbit} of equivalent orientations under the molecular point group.
Since $\mathcal{G}_i$ is nontrivial, the coarse-graining map must be set-valued. If $Q^{(i)} = \mathcal{R}(C)$ is a valid PCA frame, then for every $g \in \mathcal{G}_i$ the frame $g\,Q^{(i)}$ is equally valid, since $g$ fixes the physical configuration. The set-valued map is therefore the orbit of $Q^{(i)}$ under $\mathcal{G}_i$:
\begin{align}
    \hat{\mathcal{R}}(C) = \{g\,Q^{(i)} \mid g \in \mathcal{G}_i\} = \mathrm{Orb}_{\mathcal{G}_i}(Q^{(i)}).
\end{align}

\textbf{Claim.} The set-valued map $\hat{\mathcal{R}}$ is $SO(3)$-equivariant: $\hat{\mathcal{R}}(h \cdot C) = h\,\hat{\mathcal{R}}(C)$ for all $h \in SO(3)$.

\textbf{Proof.} The proof has two steps: first we identify the point group of a rotated molecule, then we compute the orbit.

\emph{Step 1: Conjugation of the point group.} If $g \in \mathcal{G}_i$ fixes the physical configuration of $C$, then $h\,g\,h^{-1}$ fixes that of $h \cdot C$:
\begin{align}
    (h\,g\,h^{-1}) \cdot (h \cdot C) = h\,(g \cdot C),
\end{align}
which is physically indistinguishable from $h \cdot C$ because $g \cdot C$ is physically indistinguishable from $C$. This gives an isomorphism $\mathcal{G}_i \to \mathcal{G}_{h \cdot C}$ via $g \mapsto h\,g\,h^{-1}$, so the point group of the rotated configuration is $\mathcal{G}_{h \cdot C} = h\,\mathcal{G}_i\,h^{-1}$.

\emph{Step 2: Equivariance of the orbit.} By equivariance of PCA, the frame of $h \cdot C$ is $h\,Q^{(i)}$. From Step 1, the point group of $h \cdot C$ is $h\,\mathcal{G}_i\,h^{-1}$. Applying the set-valued map to $h \cdot C$:
\begin{align}
    \hat{\mathcal{R}}(h \cdot C) &= \{g'\,h\,Q^{(i)} \mid g' \in h\,\mathcal{G}_i\,h^{-1}\} \nonumber \\
    &= \{(h\,g\,h^{-1})\,h\,Q^{(i)} \mid g \in \mathcal{G}_i\} \nonumber \\
    &= \{h\,g\,Q^{(i)} \mid g \in \mathcal{G}_i\} = h\,\hat{\mathcal{R}}(C),
\end{align}
where the first line expands the definition of $\hat{\mathcal{R}}$ using the frame and point group of $h \cdot C$, the second substitutes $g' = h\,g\,h^{-1}$, and the third cancels $h^{-1}h = I$. \hfill $\square$

We resolve the need for a multivalued coarse graining map via data augmentation described  below.

\paragraph{PCA Degeneracy} The PCA-based orientation assignment suffers from two well-known sources of degeneracy (see \cite{li_closer_2021} for a thorough review). The first is \emph{sign ambiguity}: each eigenvector $e_k$ is determined only up to a sign flip $e_k \mapsto -e_k$, giving $2^3 = 8$ possible sign assignments. However, only 4 of these preserve $\det(Q^{(i)}) > 0$, i.e.\ correspond to proper rotations in $SO(3)$; the remaining 4 produce improper rotations with $\det = -1$. The valid sign combinations (assuming that the determinant of the $+,+,+$ combination is $>0$) are
\begin{align}
    (+e_1, +e_2, +e_3), \quad (-e_1, -e_2, +e_3), \quad (+e_1, -e_2, -e_3), \quad (-e_1, +e_2, -e_3)
\end{align}
corresponding to flipping zero or two axes. The second source is \emph{order ambiguity}: permuting the three eigenvectors yields $3! = 6$ valid orderings, each defining a distinct frame. In total, this gives $4 \times 6 = 24$ ambiguities of the PCA-based canonical pose.

\paragraph{Data augmentation} More discussion of the insufficiency of plain PCA can be found in the appendix of Gao and Günnemann \cite{gao_ab-initio_2022}.  To resolve this issue we apply a simple data augmentation that spans the orbit $\hat{\mathcal{R}}(C)$ over the course of training. Given a molecular configuration $\{c^{(j)}\}_{j \in S_i}$, we perturb the atomic positions with small isotropic noise
\begin{align}
    \bar{c}^{(j)} = c^{(j)} + \epsilon^{(j)}, \qquad \epsilon^{(j)} \sim \mathcal{N}(0, \sigma^2 I_3), \qquad \sigma = 0.01\;\text{\AA}
\end{align}
and apply the coarse graining map to obtain a perturbed frame
\begin{align}
    (\bar{q}^{(i)}, \bar{Q}^{(i)}) = \mathcal{C}(\{\bar{c}^{(j)}\}_{j \in S_i})
\end{align}
The local coordinates are then computed using the perturbed frame but the original positions
\begin{align}
    \tilde{c}^{(j)} = (\bar{Q}^{(i)})^\top (c^{(j)} - q^{(i)})
\end{align}
The noise breaks the exact symmetry of the molecule, so the PCA eigenbasis is generically non-degenerate and $\bar{Q}^{(i)}$ is single-valued. Different noise realizations produce frames near different elements of the orbit $\hat{\mathcal{R}}(C)$, so over training the model sees all equivalent poses. Since the noise enters only through $\bar{Q}^{(i)}$ and not the local coordinates $\tilde{c}^{(j)}$, the body-frame geometry is preserved. This removes the need for an explicit canonicalization of the molecular orientation and allows the model to see the breadth of the orbits. 

In our setting, we sort the eigenvalues in decreasing order, which fixes the ordering and eliminates the 6 order ambiguities. To handle the residual 4 sign ambiguities, we augment each training example by randomly sampling one of the four valid sign combinations. Near-degenerate eigenvalues (which would reintroduce order ambiguity via floating point errors) are resolved by the noise perturbation described above, which generically lifts the degeneracy and ensures the eigenvalue ordering is well-defined.

\paragraph{Cell Transformations} As mentioned previously, the lattice vectors of a crystal are not unique. 
Two sets of lattice vectors $L$ and $\bar{L}$ span the same lattice if and only if \cite{arndt_efficient_2009}
\begin{align}
    \bar{L} = ML, \qquad M \in GL(3,\mathbb{Z}), \quad |\det M| = 1
\end{align}
where $GL(3,\mathbb{Z})$ is the group of $3\times 3$ integer matrices with determinant $\pm 1$ (unimodular matrices). 
Under this transformation the fractional coordinates transform as $f^{(i)} \mapsto f^{(i)} M^{-1}$ so that the Cartesian positions $c^{(i)} = f^{(i)}L$ are unchanged. The molecular orientations $Q^{(i)}$ are similarly unaffected. More generally, an integer matrix $M$ with $|\det M| = m > 1$ produces a supercell containing $m$ copies of the original unit cell \cite{jin_oxtal_2025}.
In this work we neglect invariance to both unimodular basis changes and supercell equivalences, training on a single cell choice present in the dataset. 
This has been effective in practice for inorganic crystal structure prediction, and we leave explicit treatment of these symmetries to future work.
We note that a cluster-based description avoids the need to account for this because it has no lattice and uses Cartesian coordinates. 

\paragraph{Permutations}
A crystal is invariant under permutation of molecule indices $i = 1, \ldots, M$ and atom indices $j = 1, \ldots, N$:
\begin{align}
    \Phi_g: U \to U, \quad P \to P, \quad f^{(\sigma(i))} \to f^{(i)}, \quad Q^{(\sigma(i))} \to Q^{(i)}
\end{align}
for $\sigma \in S_M$. We handle this symmetry by choosing our network to be permutation invariant with respect to both molecule and atom reorderings. See Section~\ref{app:architecture} for details.

\paragraph{Space groups and impact of molecular coarse-graining}

The space group $\mathcal{G}$ of a molecular crystal structure is the group of all Seitz operations $\{R \mid \mathbf{t}\}$ that map the crystal to itself while preserving
atomic types:
\begin{equation}
    \mathcal{G} = \bigl\{\{R \mid \mathbf{t}\} \in E(3) \;\big|\;
        \{R \mid \mathbf{t}\} \cdot \{c^{(j)}, a^{(j)}\}_{j=1}^N
        = \{c^{(j)}, a^{(j)}\}_{j=1}^N \bigr\}.
\end{equation}
The space group of the coarse-grained molecular crystal is the analogous stabilizer acting on
the CG descriptors:
\begin{equation}
    \mathcal{G}_{\mathrm{CG}} = \bigl\{\{R \mid \mathbf{t}\} \in E(3) \;\big|\;
        \{R \mid \mathbf{t}\} \cdot \{q^{(i)}, Q^{(i)}\}_{i=1}^M
        = \{q^{(i)}, Q^{(i)}\}_{i=1}^M \bigr\}.
\end{equation}

We suspect that coarse-graining can thus only ``increase'' the space group symmetry or leave it unchanged: $\mathcal{G} \leq \mathcal{G}_{\mathrm{CG}}$.
The mechanism is geometric: molecular centroids tend to occupy high-symmetry packing positions. 
This empirical observation is a simpler version of those made in CrystalMath~\cite{galanakis_rapid_2024},
and the CG descriptor $(q^{(i)}, Q^{(i)})$ is a low-resolution summary that retains less information.
It is the molecular shape that breaks the higher
$\mathcal{G}_{\mathrm{CG}}$ symmetry down to $ \mathcal{G}$. 

We do not explicitly enforce space group symmetry in the generative model. Instead, we follow the common approach in crystal structure prediction of learning in $P1$ (the trivial space group with no non-trivial symmetry operations) and relying on the training data distribution to implicitly capture the statistics of higher-symmetry structures. The coarse-graining further simplifies this: since $\mathcal{G} \leq \mathcal{G}_{\mathrm{CG}}$, the CG representation is at least as symmetric as the atomistic one, and a model that generates valid CG packings will tend to respect the dominant space group motifs present in the data. The fine-grained space group $\mathcal{G}$ is then recovered upon reconstruction of the full atomistic structure from the CG descriptors and the stored local coordinates $\tilde{c}^{(j)}$.

\section{Neural Network Architecture} \label{app:architecture}

\begin{figure}[H]
    \centering
    \includegraphics[width=0.75\linewidth]{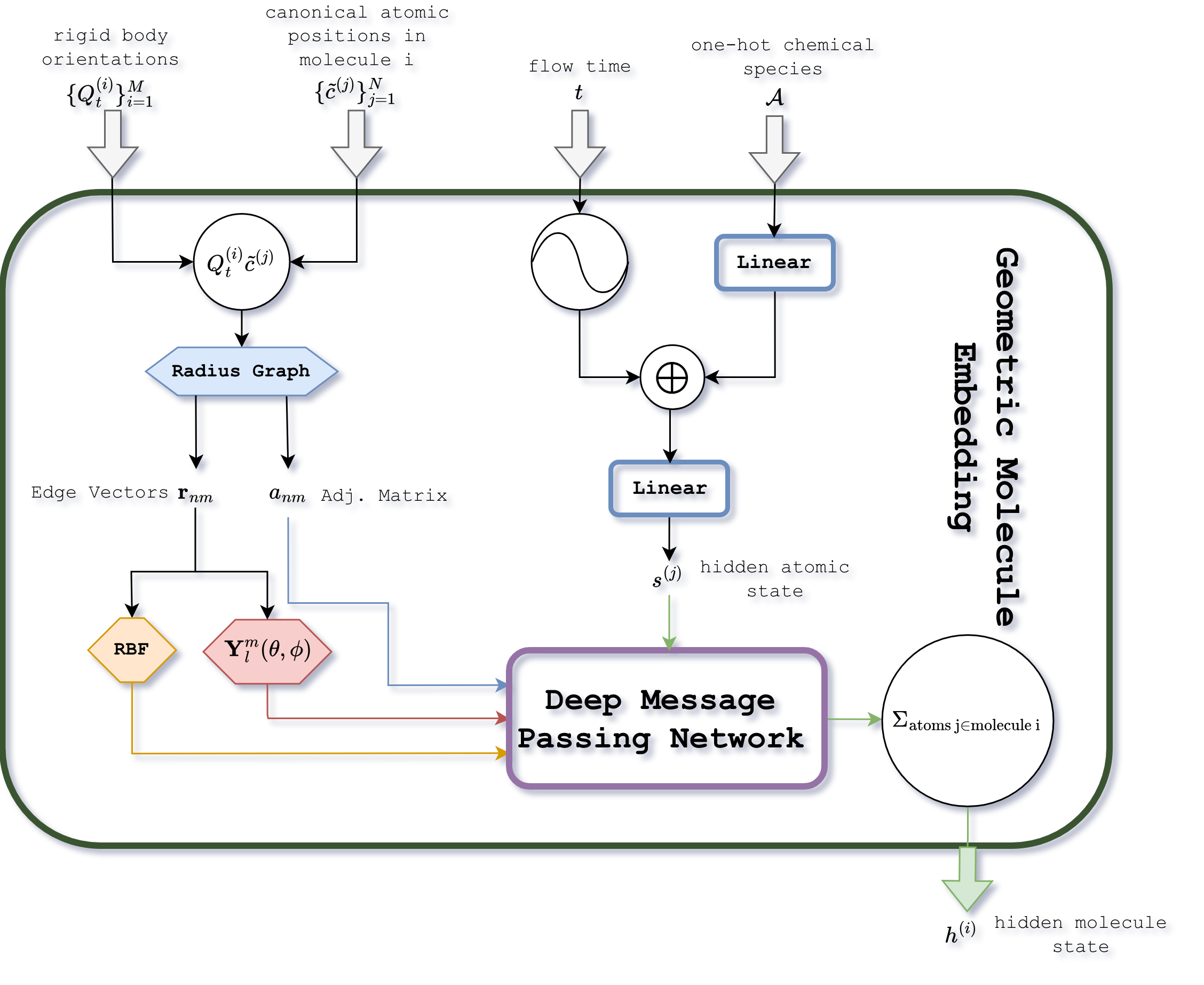}
    \caption{\textbf{Geometric molecule embedding.} Rotated atomic coordinates define a radius graph; edge lengths are expanded with a Gaussian radial basis and edge directions with spherical harmonics, then combined with species and time embeddings to form equivariant node features to be fed into a deep message passing network whose final hidden states are averaged to produce a molecule embedding.}
    \label{fig:geometric_embed}
\end{figure}

\paragraph{Geometric Molecule Embedding} The CG-OMatG network operates in two stages. In the first stage, it constructs a molecule embedding via the
\texttt{Geometric Molecule Embedding} module (Figure~\ref{fig:geometric_embed}). The module takes as input the tuple
\begin{align}
    (\{Q_t^{(i)}\}_{i=1}^M, \{\tilde{c}^{(j)}\}_{j=1}^N, t, \mathcal{A}).
\end{align}
Each canonical atomic coordinate is rotated by the time-dependent rotation matrices:
\begin{align}
    c^{(j)} = Q_t^{(i)} \tilde{c}^{(j)} + q^{(i)}. \label{eq:rotinput}
\end{align}
This produces a time-dependent atomic position.

A radius graph is then built using the rotated coordinates. Two atoms $m$ and $n$ are connected if
\begin{align}
    \|c^{(n)} - c^{(m)}\| \le r_{\text{cut}} \quad  \text{ and }  \quad n,m\in S_i
\end{align}
in which case the adjacency matrix has entry $a_{nm}=1$. This means that the atoms must be nearby and within the same molecule. For each edge $e_k$ (with $k=1,\dots,E$), we form radial
features by expanding the edge length $\|e_k\|$ in a set of radial basis functions using \texttt{e3nn}'s
\texttt{soft\_one\_hot\_linspace}. This can be viewed as a projection onto a basis:
\begin{align}
    y_l(e_k) = Z^{-1} f_l(\|e_k\|),
    \qquad \text{with} \qquad
    \left\langle \sum_{l=1}^{l_\text{max}} y_l(e_k)^2 \right\rangle_{e_k} \approx 1,
\end{align}
where $\langle \cdot \rangle_{e_k}$ denotes an average over edges.

In this work, at the intramolecular message passing stage, we use the Gaussian basis with \texttt{cutoff=True}. Let $l_\text{max}$ denote the number of radial basis functions and define the spacing and centers (excluding endpoints) by
\begin{align}
    \Delta = \frac{r_{\text{cut}}}{l_\text{max}+1},
    \qquad
    c_l = l \Delta,
    \qquad \ell=1,\dots,l_\text{max}.
\end{align}
Then the $l$th radial basis component is
\begin{align}
    y_l(e_k) = \frac{1}{1.12}\exp\left(-\left(\frac{\|e_k\|-c_l}{\Delta}\right)^2\right)
\end{align}  
For more details, see the \texttt{e3nn} documentation.

In addition to radial features, we compute angular features by applying spherical harmonics to the normalized edge directions $e_k/\|e_k\|$. For each $\ell$, the spherical harmonics define a map $Y^\ell:\mathbb{R}^3\to \mathbb{R}^{2\ell+1}$ satisfying rotation equivariance:
\begin{align}
    Y^\ell(Rx) = D^\ell(R)Y^\ell(x),
\end{align}
where $D^\ell(R)$ is the Wigner-$D$ matrix for rank-$\ell$ irreducible representations. We normalize them such that
$\|Y^\ell(x)\| = 2\ell+1$. By equivariance, applying the time-dependent rotations $Q_t^{(i)}$ in \eqref{eq:rotinput} corresponds to rotating the spherical-harmonic features by the same transformation.

After embedding positional information, we embed the atomic species $\mathcal{A}$ using a learned lookup, which is
equivalent to applying a linear layer without bias to a one-hot encoding. The flow time $t$ is embedded via a
sinusoidal time embedding. All non-positional features are concatenated and passed through a linear layer to obtain
a consistent feature shape across tensor ranks. The resulting per-atom features $s^{(j)}$ contain irreducible
components of ranks $\ell=0,1,\dots,\ell_{\max}$, with $C$ channels per rank. These features are then processed by the
deep message passing network, to be described later. To obtain a molecule-level hidden state, we sum the final per-atom geometric features across all atoms in the molecule:
\begin{align}
    h^{(i)} = \sum_{j \in \text{molecule i}}^{N} s^{(j)},
\end{align}
where $s^{(j)}$ denotes the final geometric hidden feature of atom $j$ and $N_i$ is the number of atoms in molecule $i$.

\begin{figure}[t]
    \centering
    \includegraphics[width=1\linewidth]{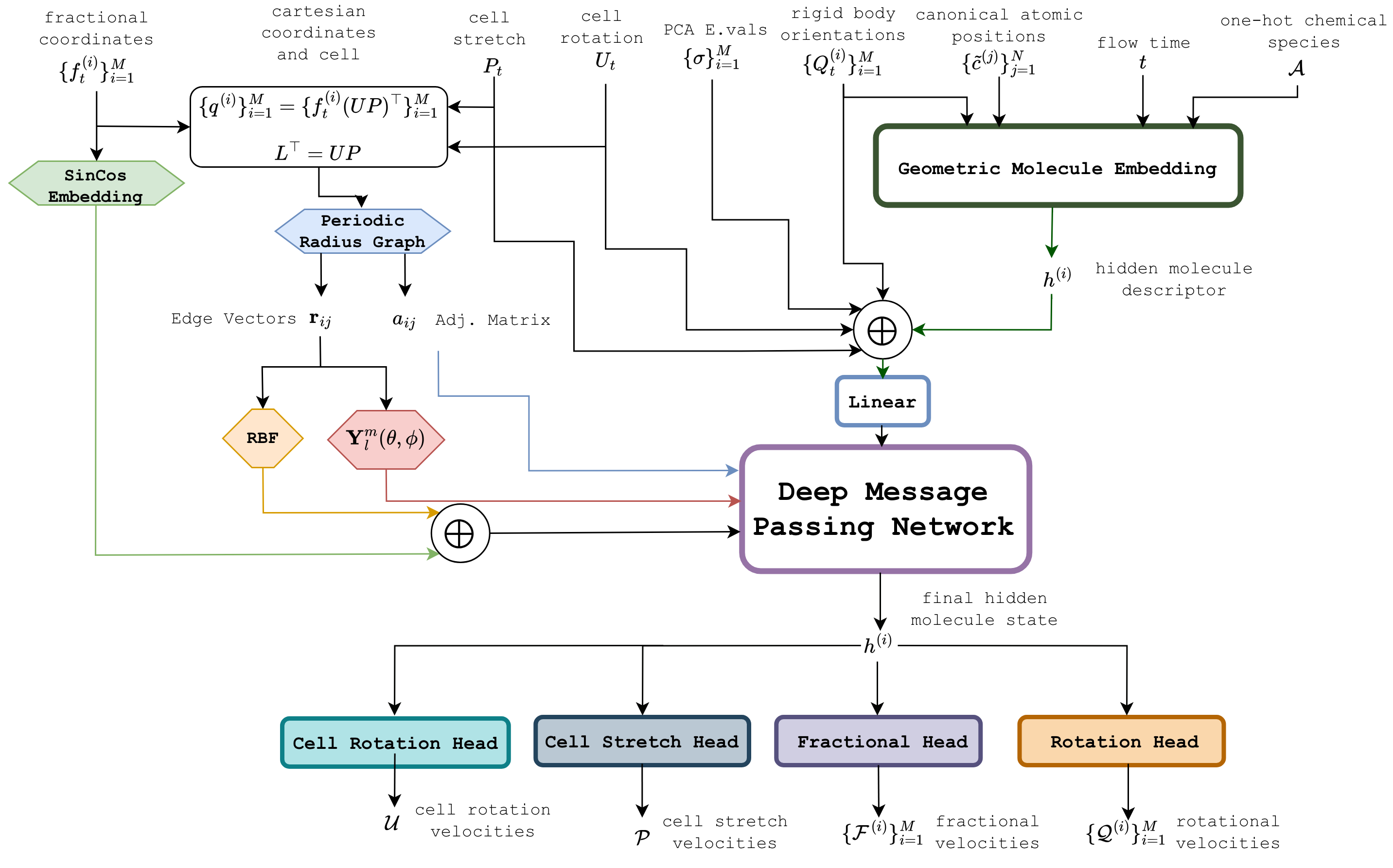}
    \caption{\textbf{Overall architecture.} Geometric inputs are embedded and propagated on a periodic radius graph over centroid coordinates (via ghost centroids), yielding per-centroid hidden states for lattice, fractional-coordinate, and rotational-velocity prediction.}
    \label{fig:CG-OMatGNet}
\end{figure}

\paragraph{Overall Architecture} The overall architecture (Figure~\ref{fig:CG-OMatGNet}) closely mirrors the molecule embedding module, but it acts on a different set of inputs. Rather than using sinusoidal time embeddings and chemical species, it uses geometric features derived from the time-dependent rotations and the time-dependent unit cell, the PCA eigenvalues of the rigid body, together with the geometric molecule embedding; these features are passed through a linear layer to obtain a consistent shape across spherical tensor ranks (meaning they share the same number of channels for all ranks). Message passing is then performed on a \emph{periodic} radius graph constructed from centroid coordinates (not the fractional coordinates). We implement periodicity by duplicating the structure to create ghost centroids, with enough replicas so that every centroid in the fundamental cell can access all neighbors within the cutoff radius $r_{\text{cut}}$. This is then fed into a regular radius graph afterwards. The resulting geometric graph is then processed by a deep message passing network.

\begin{figure}[t]
    \centering
    \includegraphics[width=0.65\linewidth]{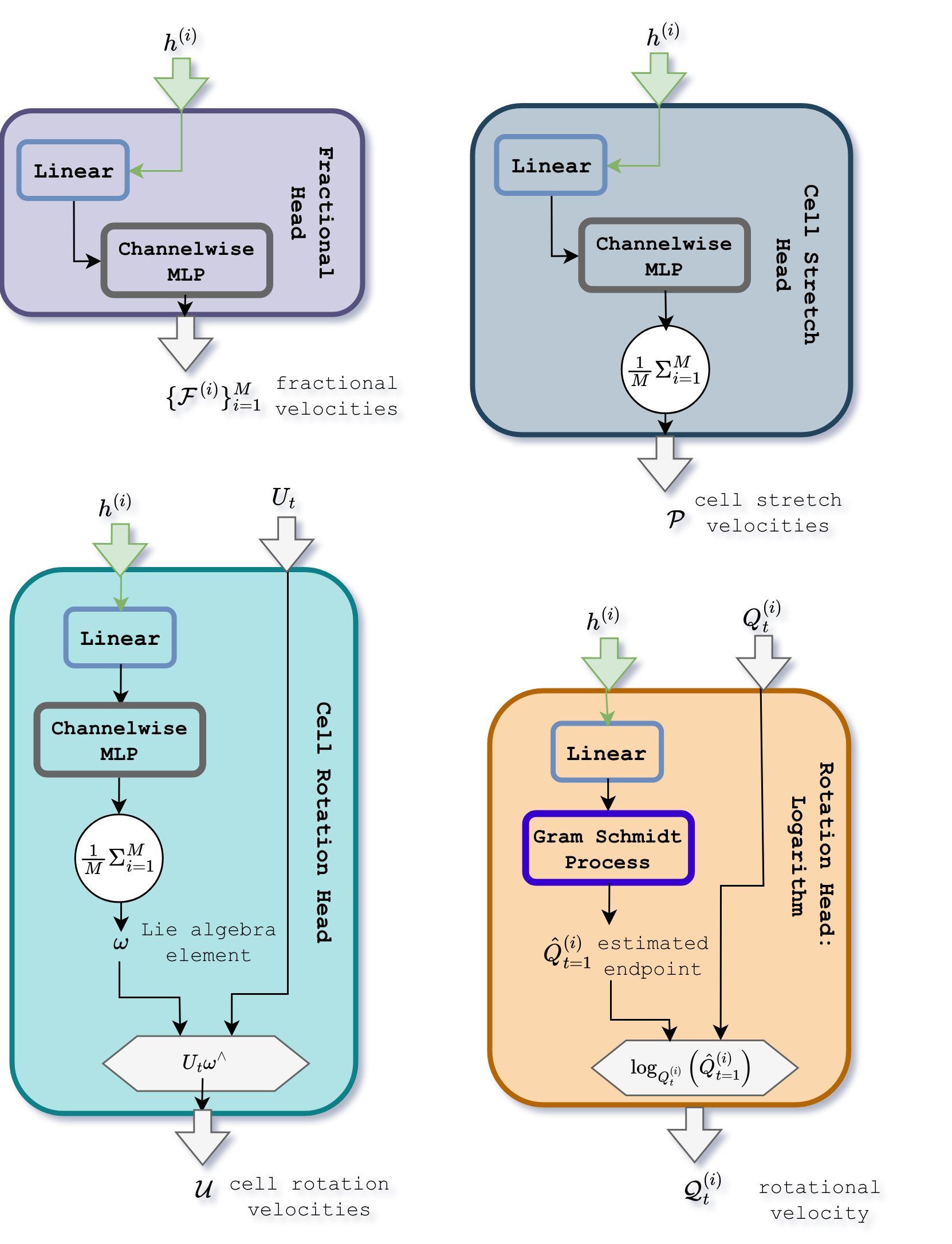} \caption{\textbf{Prediction heads.} Per-centroid hidden states feed four heads: fractional translation, cell rotation, cell stretch, and molecular orientation. Outputs are mapped to tangent vectors on each factor of $\mathcal{M}$ via Riemannian logarithm or Lie-algebra left-translation.}      
  \label{fig:CG-OMatG-heads}
\end{figure}

\paragraph{Prediction Heads}
The prediction heads act on the per-centroid hidden state at the output of the crystal branch and emit one tangent vector per modeled field of $\mathcal{M}$.  The fractional-position head outputs a per-centroid translational tangent vector $\mathcal{F}^{(i)} \in T_{f^{(i)}}\mathbb{T}^3 \cong \mathbb{R}^3$. The cell is decomposed as $L_t = (U_t P_t)^{\!\top}$ with $U_t \in SO(3)$ and $P_t \in \mathrm{Sym}_3^+$, and is modeled by two heads.  The cell-rotation head reads out $\omega_U^{(i)} \in \mathbb{R}^3$ at every centroid, mean-pools over the centroids in the unit cell to a single $\omega_U \in \mathfrak{so}(3) = T_I SO(3)$ per cell, and lifts it to $T_{U_t}SO(3)$ by left translation,
\begin{align}
    \mathcal{U}_t = U_t\,\omega_U^{\wedge}.
\end{align}
The cell-stretch head reads out the six independent Voigt components of a
symmetric $3\!\times\!3$ matrix at every centroid, mean-pools over the
centroids, and reshapes to a tangent vector
$\mathcal{P}_t \in T_{P_t}\mathrm{Sym}_3^+ = \mathrm{Sym}_3$; the full cell
velocity follows from the product rule,
\begin{align}
    \dot L_t = \big(\mathcal{U}_t P_t + U_t \mathcal{P}_t\big)^{\!\top}.
\end{align}
For the per-molecule rotations $Q^{(i)}_t$, we consider two interchangeable
heads that both produce a tangent vector $\mathcal{Q}^{(i)}_t \in T_{Q^{(i)}_t}SO(3)$.  The first predicts a denoised endpoint $\hat Q^{(i)}_{t=1}$ and maps it back via the Riemannian logarithm,
\begin{align}
   \mathcal{Q}^{(i)}_t = \log_{Q^{(i)}_t}\!\big(\hat Q^{(i)}_{t=1}\big).
\end{align}
To enforce $\hat Q^{(i)}_{t=1} \in SO(3)$ the head outputs two
$\ell\!=\!1$ vectors which are orthonormalized via Gram--Schmidt, with their cross product completing the rotation matrix; molecules whose two predicted vectors are nearly collinear are masked out of the rotation
loss. 

\begin{figure}[t]
    \centering
    \includegraphics[width=1\linewidth]{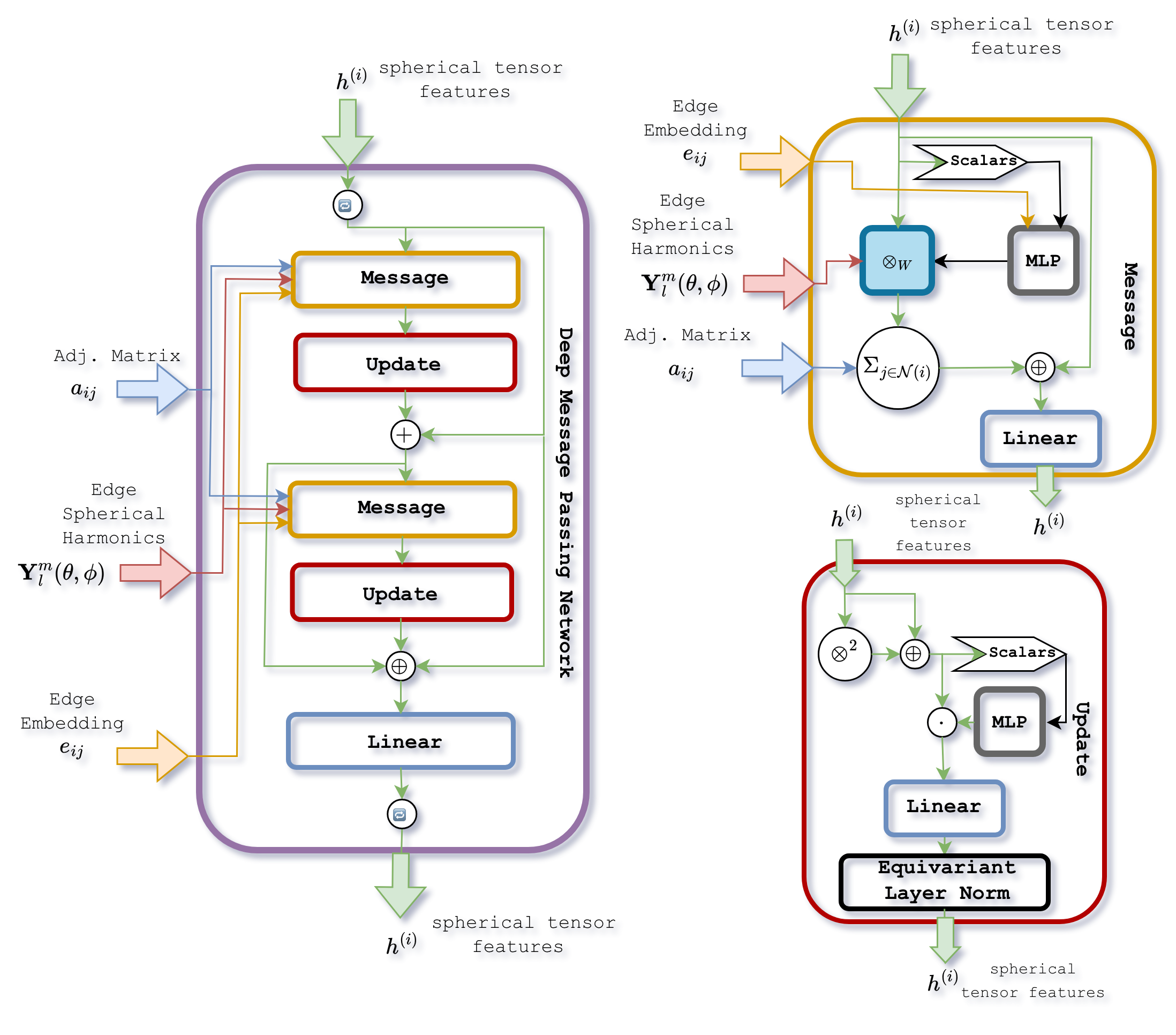}
    \caption{\textbf{Deep message-passing network.} Left: two message-update blocks with a residual-sum skip and a final concatenation skip. Right: the message layer uses an RBF-conditioned weighted tensor product with spherical harmonics; the update layer applies tensor augmentation, scalar gating, and geometric layer normalization.}
    \label{fig:deepmessage}
\end{figure}

\paragraph{Deep Message Passing Network} Our message-passing network closely follows EquiJump~\cite{costa_equijump_2024}; see that work for further details. For completeness, we describe it here. Each deep message-passing network is built from a repeated sequence of message-update blocks with two skip connections. Starting from state (a), a message-update block produces an intermediate representation (b) (the node state immediately before the first $+$ in the diagram). This intermediate is combined with (a) through a residual summation to yield the post-skip state (b) (immediately after the $+$). A second message-update block is then applied to produce the current state (the node state immediately before $\oplus$). Finally, a concatenation skip forms $[\text{(a)},\text{(b)},\text{current}]$, which is projected back to the hidden dimension by a linear layer. This full pattern is repeated some number of times (as indicated by the repeat symbol), with independent parameters in each repetition. The overall block structure is shown on the left of Figure~\ref{fig:deepmessage}.

In the message layer, each node's features are split into a scalar stream and a spherical-tensor stream. The scalar stream is concatenated with the edge radial basis embedding and passed through an MLP to produce mixing weights. These weights parameterize a weighted tensor product between the node's spherical-tensor features and the edge spherical harmonics, yielding edge messages. Messages are summed over each node's neighborhood, concatenated with the node's original state, and mapped back to the hidden dimension with a linear layer. In the update layer, tensor features are augmented by concatenating the tensor-square with the original tensor features. In parallel, the scalar stream is split to produce scalar gates that multiplicatively modulate the tensor features. The resulting tensor features are then passed through a linear layer followed by geometric layer normalization \cite{liao_equiformer_2023}. The message and update operations are shown on the right of Figure~\ref{fig:deepmessage}.

\section{UMA Relaxation Scheme}\label{app:relax}

For every candidate molecular crystal, we evaluate the \texttt{uma-s-1p2} universal foundation potential energy model~\cite{wood_uma_2026} via its \texttt{FAIRChemCalculator} ASE interface, using the molecular-crystal task head trained on the OMC25 dataset~\cite{gharakhanyan_open_2025}. The structure is then relaxed under a three-stage BFGS protocol implemented in ASE~\cite{hjorth_larsen_atomic_2017} that mirrors the relaxation pipeline of MolCrystalFlow~\cite{zeng_molcrystalflow_2026}:

\begin{enumerate}[leftmargin=*]
  \item \emph{Rigid-body warm-up.} Each molecular building block is held
        internally rigid by a custom ASE constraint: at every BFGS step,
        positions are re-projected onto the rigid-body manifold by
        Kabsch alignment to the block's reference geometry, and per-atom
        forces are replaced by the corresponding net-force / net-torque
        contributions of that block. We run BFGS for
        $N_{\text{rigid}}=100$ iterations, allowing centroids and
        orientations to relax while intramolecular geometries are
        preserved. 
  \item \emph{Coupled cell + atomic relax.} Lattice and atomic degrees
        of freedom are co-optimised by wrapping the system in an ASE
        \texttt{FrechetCellFilter}, whose
        coordinates are the atomic positions in the undeformed cell
        together with the matrix logarithm of the deformation gradient.
        BFGS is run on the filtered system until the maximum atomic
        force component falls below $f_{\max}=0.01\,\mathrm{eV}/$\AA
        or a $1000$-step cap is reached.
  \item \emph{Atomic-only relax.} The cell is then fixed and a final
        BFGS pass on the atomic coordinates re-converges them to the
        same $f_{\max}=0.01\,\mathrm{eV/}$\AA tolerance, eliminating any
        residual atomic forces left over from the previous stage.
\end{enumerate}

For each structure we record the initial and relaxed UMA energies, the relaxation gain $\Delta E = E_{\text{init}} - E_{\text{relaxed}}$, the input-to-final atomic RMSD, the maximum lattice-vector drift, per-stage step counts, and convergence flags (a stage is convergent iff it exits before the step cap, i.e.\ on the $f_{\max}$ criterion). Per-step trajectories of all three stages can optionally be stitched together
for inspection.

\section{Hyperparameters} \label{app:hyperparams}
\begin{table}[H]
  \centering
  \caption{MolCrystalFlow inference hyperparameters used with the
    OMC25-MCF checkpoint, following the values recommended in the
    project README.}
  \label{tab:mcf-inference-hparams}
  \small
  \begin{tabular}{@{}ll@{}}
    \toprule
    \textbf{Parameter} & \textbf{Value} \\
    \midrule
    \multicolumn{2}{@{}l}{\textit{Inference}} \\[2pt]
    \quad \texttt{config-name}    & \texttt{omc25\_inference.yaml} \\
    \quad \texttt{ckpt\_path}     & \texttt{model-checkpoints/omc25-mcf/best.ckpt} \\
    \quad \texttt{num\_samples}   & 30 \\[4pt]
    \multicolumn{2}{@{}l}{\textit{Interpolant — sampling}} \\[2pt]
    \quad \texttt{num\_timesteps} & 50 \\[4pt]
    \multicolumn{2}{@{}l}{\textit{Interpolant — translations}} \\[2pt]
    \quad \texttt{scaling}        & 9.0\,\AA \\[4pt]
    \multicolumn{2}{@{}l}{\textit{Interpolant — rotations}} \\[2pt]
    \quad \texttt{exp\_rate}      & 3.0 \\[4pt]
    \multicolumn{2}{@{}l}{\textit{Model — backbone embedder}} \\[2pt]
    \quad \texttt{num\_atom\_types} & 12 \\[4pt]
    \multicolumn{2}{@{}l}{\textit{Aggregation}} \\[2pt]
    \quad Draws per crystal $K$   & 30 \\
    \quad Sampling strategy       & Independent draws from the flow ODE \\
    \bottomrule
  \end{tabular}
\end{table}

\begin{table}[H]
  \centering
  \caption{Genarris hyperparameters used for the molecular-crystal
    generation baseline.}
  \label{tab:genarris-hparams}
  \small
  \begin{tabular}{@{}ll@{}}
    \toprule
    \textbf{Parameter} & \textbf{Value} \\
    \midrule
    \multicolumn{2}{@{}l}{\textit{Master}} \\[2pt]
    \quad \texttt{Z}                          & Per-crystal (\#unique \texttt{bb\_indices}) \\
    \quad MPI ranks                           & 8 \\[4pt]
    \multicolumn{2}{@{}l}{\textit{Workflow}} \\[2pt]
    \quad \texttt{tasks}                      & \texttt{generation}, \texttt{symm\_rigid\_press} \\[4pt]
    \multicolumn{2}{@{}l}{\textit{Generation}} \\[2pt]
    \quad \texttt{generation\_type}           & \texttt{crystal} \\
    \quad \texttt{spg\_distribution\_type}    & \texttt{standard} \\
    \quad \texttt{num\_structures\_per\_spg}  & 30 \\
    \quad \texttt{unit\_cell\_volume\_mean}   & \texttt{predict} \\
    \quad \texttt{volume\_mult}               & 1.5 \\
    \quad \texttt{sr}                         & 0.85 \\
    \quad \texttt{natural\_cutoff\_mult}      & 1.2 \\
    \quad \texttt{tol}                        & 0.01 \\
    \quad \texttt{max\_attempts\_per\_spg}    & $10^{7}$ \\
    \quad \texttt{max\_attempts\_per\_volume} & $10^{7}$ \\[4pt]
    \multicolumn{2}{@{}l}{\textit{Symmetric rigid-body relaxation}} \\[2pt]
    \quad \texttt{method}                     & BFGS \\
    \quad \texttt{sr}                         & 0.85 \\
    \quad \texttt{natural\_cutoff\_mult}      & 1.2 \\
    \quad \texttt{tol}                        & 0.01 \\[4pt]
    \multicolumn{2}{@{}l}{\textit{Aggregation}} \\[2pt]
    \quad Draws per crystal $K$               & 30 \\
    \quad Sampling strategy                   & W/o replacement; w/ replacement if $\#\text{converged} < K$ \\
    \bottomrule
  \end{tabular}
\end{table}

\begin{table}[H]
\centering
\small
\caption{CG-OMatG pre-training hyperparameters. Each MLP has a single hidden layer with the listed width.}
\label{tab:hparams}
\begin{tabular}{@{}p{0.38\linewidth}p{0.55\linewidth}@{}}
\toprule
\textbf{Parameter} & \textbf{Value} \\
\midrule
\multicolumn{2}{@{}l}{\textbf{Training}}\\[2pt]
\quad Batch size (global)
  & $4 \times 320 = 1280$ \\
\quad Optimizer
  & AdamW, lr $= 5\!\times\!10^{-4}$, weight decay $= 0.00818$ \\
\quad LR schedule
  & Cosine annealing, 1500 epochs, $\eta_{\min} = 10^{-7}$ \\
\quad Gradient clipping
  & 0.5, per-element \\
\quad Time sampling
  & Logit-normal: $t = \sigma(1.7\,Z + 0.8)$, $Z \sim \mathcal{N}(0,1)$ \\
\quad PCA-frame augmentation scale
  & Isotropic Gaussian noise ($\sigma = 0.01$\,\text{\AA}) \\[5pt]

\multicolumn{2}{@{}l}{\textbf{Molecule branch} --- intra-molecule message passing}\\[2pt]
\quad Layers / channels / irrep rank
  & 1 / 16 / 1 \\
\quad Edge construction
  & Spherical harmonics $\ell_{\max}\!=\!2$, 64 Gaussian radial bases, cutoff $r_c = 8.0$\,\text{\AA}, max 100 neighbors \\
\quad Edge-weight / node-update MLPs
  & [512] / [512] \\
\quad Species embedding
  & Dimension 64, vocabulary size 100 \\
\quad Atom $\to$ molecule pooling
  & Sum reduction \\[5pt]

\multicolumn{2}{@{}l}{\textbf{Crystal branch} --- inter-molecular message passing}\\[2pt]
\quad Layers / channels / irrep rank
  & 5 / 16 / 2 \\
\quad Edge construction
  & Spherical harmonics $\ell_{\max}\!=\!2$, 64 Gaussian radial bases + 64-frequency sin/cos Fourier features on fractional BB--BB displacements \\
\quad Time-conditioned cutoff $r_c(t)$
  & Lagrange interpolation through $(t,\,r_c) = \{(0,\,9.0),\;(0.5,\,10.0),\;(1,\,11.5)\}$\,\text{\AA}; max 100 neighbors \\
\quad Edge-weight / node-update MLPs
  & [1024] / [256] \\
\quad Rotation readout
  & Gram--Schmidt orthogonalization + $\mathrm{Log}_{SO(3)}$ \\[5pt]

\multicolumn{2}{@{}l}{\textbf{Flow matching} --- per-field interpolants}\\[2pt]
\quad $\mathbb{T}^3$ fractional positions
  & Periodic linear interpolant, Euler ODE; center-of-mass motion subtracted before loss \\
\quad $SO(3)$ molecule orientations
  & Riemannian geodesic interpolant, MODE \\
\quad $\mathcal{S}\!ym_3^+(\mathbb{R})$ lattice shape
  & Riemannian geodesic interpolant on SPD manifold ($\mathrm{Exp}$/$\mathrm{Log}$/$\langle\cdot,\cdot\rangle_{\mathrm{SPD}}$), MODE \\
\quad $SO(3)$ cell orientation
  & Riemannian geodesic interpolant, MODE \\[5pt]

\multicolumn{2}{@{}l}{\textbf{Loss weights} ($\mathbb{T}^3$ / $SO(3)_{\mathrm{mol}}$ / $\mathcal{S}\!ym_3^+$ / $SO(3)_{\mathrm{cell}}$)}\\[2pt]
\quad
  & 15.0 / 3.0 / 1.0 / 1.0 \\[5pt]

\multicolumn{2}{@{}l}{\textbf{Priors} ($t = 0$)}\\[2pt]
\quad Fractional positions
  & Uniform on $[0,1)^3$ \\
\quad Orientations (mol.\ + cell)
  & Haar-uniform on $SO(3)$ \\
\quad Lattice
  & Log-normal lengths + uniform angles on $[60^\circ,120^\circ]$, parameters fit from CSD \\[5pt]

\multicolumn{2}{@{}l}{\textbf{Inference}}\\[2pt]
\quad ODE integration
  & 500 Euler steps \\
\quad Velocity annealing
  & $\hat{b}(t) = (1 + \alpha_v\,t)\,b(t)$; $\alpha_v$: positions 8.0, mol.\ orient.\ 8.0, cell rot.\ 0.1, lattice shape 0.0 \\
\bottomrule
\end{tabular}
\end{table}

\begin{table}[H]
\centering
\small
\caption{CG-OMatG RL fine-tuning hyperparameters (GRPO with PPO clipping).}
\label{tab:rl_hparams}
\begin{tabular}{@{}p{0.38\linewidth}p{0.55\linewidth}@{}}
\toprule
\textbf{Parameter} & \textbf{Value} \\
\midrule
\multicolumn{2}{@{}l}{\textbf{Training}}\\[2pt]
\quad Optimizer
  & Adam, lr $= 10^{-4}$ \\
\quad Max steps
  & 5000 \\
\quad Gradient clipping
  & 1.0, global norm \\[5pt]

\multicolumn{2}{@{}l}{\textbf{GRPO / PPO}}\\[2pt]
\quad Group size / num.\ groups
  & 64 / 5 \\
\quad Shared $x_0$ within group
  & Yes \\
\quad PPO clip $\epsilon$
  & 0.1 \\
\quad PPO epochs per step
  & 1 \\[5pt]

\multicolumn{2}{@{}l}{\textbf{Exploration noise} ($\sigma(t) = \sigma_0 \sqrt{t}$)}\\[2pt]
\quad $\mathbb{T}^3$ / $SO(3)_{\mathrm{mol}}$ / $\mathcal{S}\!ym_3^+$
  & 0.1 / 0.1 / 0.1 \\
\quad $SO(3)_{\mathrm{cell}}$
  & 0.01 \\[5pt]

\multicolumn{2}{@{}l}{\textbf{Policy loss weights} (pos / mol.\ rot / lattice / cell rot)}\\[2pt]
\quad Policy
  & 1.0 / 1.0 / 0.5 / 0.25 \\
\quad KL regularization
  & $10^{-3}$ (all fields) \\[5pt]

\multicolumn{2}{@{}l}{\textbf{Reward} (UMA energy)}\\[2pt]
\quad Scale
  & 1.0 \\
\quad Invalid-structure penalty
  & 3.0 eV/atom \\
\quad Volume-check cutoff
  & 0.1 \\
\quad Polar-sine cutoff
  & 0.001 \\
\bottomrule
\end{tabular}
\end{table}

\clearpage
\section{Additional Results}\label{app:additional_results}

\begin{table}[H]
  \centering
  \scriptsize
  \caption{CCDC packing-similarity metrics for the three homomolecular sixth CSD blind-test targets ($k=30$ inference). Values are rates; $\uparrow$~higher is better and $\downarrow$~lower is better.}
  \label{tab:bt6_results}
  \setlength{\tabcolsep}{3.5pt}
  \resizebox{\textwidth}{!}{%
  \begin{tabular}{@{}llccccc@{}}
    \toprule
    Target & Method & Solved $\uparrow$ & Solved (collisions allowed) $\uparrow$ & Packing match $\uparrow$ & Packing match (per draw) $\uparrow$ & Clash $\downarrow$ \\
    \midrule
    NACJAF & OXtal & $0.30\pm0.15$ & $0.30\pm0.15$ & $\boldsymbol{1.00\pm0.00}$ & $\boldsymbol{0.08\pm0.01}$ & $\boldsymbol{0.00\pm0.00}$ \\
            & MCF & $0.00$ & $0.00$ & $0.00$ & $0.00$ & $0.93$ \\
            & CG-OMatG & $0.10\pm0.10$ & $0.10\pm0.10$ & $0.40\pm0.16$ & $0.02\pm0.01$ & $0.20\pm0.02$ \\
            & CG-OMatG-IRL & $0.00\pm0.00$ & $0.00\pm0.00$ & $0.50\pm0.17$ & $0.03\pm0.01$ & $0.07\pm0.01$ \\
            & CG-OMatG (relaxed) & $\boldsymbol{0.60\pm0.16}$ & $\boldsymbol{0.60\pm0.16}$ & $0.80\pm0.13$ & $0.05\pm0.01$ & $0.02\pm0.01$ \\
            & CG-OMatG-IRL (relaxed) & $0.10\pm0.10$ & $0.10\pm0.10$ & $0.50\pm0.17$ & $0.02\pm0.01$ & $\boldsymbol{0.00\pm0.00}$ \\
    \addlinespace
    XAFPAY & OXtal & $0.00\pm0.00$ & $0.00\pm0.00$ & $\boldsymbol{1.00\pm0.00}$ & $\boldsymbol{0.09\pm0.01}$ & $\boldsymbol{0.00\pm0.00}$ \\
            & MCF & $0.00$ & $0.00$ & $0.00$ & $0.00$ & $0.93$ \\
            & CG-OMatG & $0.00\pm0.00$ & $0.00\pm0.00$ & $0.10\pm0.10$ & $0.00\pm0.00$ & $0.51\pm0.02$ \\
            & CG-OMatG-IRL & $0.00\pm0.00$ & $0.00\pm0.00$ & $0.30\pm0.15$ & $0.01\pm0.01$ & $0.41\pm0.03$ \\
            & CG-OMatG (relaxed) & $\boldsymbol{0.10\pm0.10}$ & $\boldsymbol{0.10\pm0.10}$ & $0.10\pm0.10$ & $0.00\pm0.00$ & $\boldsymbol{0.00\pm0.00}$ \\
            & CG-OMatG-IRL (relaxed) & $0.00\pm0.00$ & $0.00\pm0.00$ & $0.00\pm0.00$ & $0.00\pm0.00$ & $\boldsymbol{0.00\pm0.00}$ \\
    \addlinespace
    XAFQIH & OXtal & $\boldsymbol{0.00\pm0.00}$ & $\boldsymbol{0.00\pm0.00}$ & $0.30\pm0.15$ & $\boldsymbol{0.01\pm0.01}$ & $\boldsymbol{0.00\pm0.00}$ \\
            & MCF & $\mathbf{0.00}$ & $\mathbf{0.00}$ & $0.00$ & $0.00$ & $0.73$ \\
            & CG-OMatG & $\boldsymbol{0.00\pm0.00}$ & $\boldsymbol{0.00\pm0.00}$ & $0.30\pm0.15$ & $\boldsymbol{0.01\pm0.01}$ & $0.55\pm0.03$ \\
            & CG-OMatG-IRL & $\boldsymbol{0.00\pm0.00}$ & $\boldsymbol{0.00\pm0.00}$ & $0.00\pm0.00$ & $0.00\pm0.00$ & $0.40\pm0.02$ \\
            & CG-OMatG (relaxed) & $\boldsymbol{0.00\pm0.00}$ & $\boldsymbol{0.00\pm0.00}$ & $\boldsymbol{0.40\pm0.16}$ & $\boldsymbol{0.01\pm0.01}$ & $\boldsymbol{0.00\pm0.00}$ \\
            & CG-OMatG-IRL (relaxed) & $\boldsymbol{0.00\pm0.00}$ & $\boldsymbol{0.00\pm0.00}$ & $0.30\pm0.15$ & $\boldsymbol{0.01\pm0.01}$ & $\boldsymbol{0.00\pm0.00}$ \\
    \bottomrule
  \end{tabular}}
  \vspace{0.3em}

  \parbox{0.98\textwidth}{\footnotesize Error bars are SEM over ten blocks of 30 draws. MCF has one $K=30$ block and therefore no error bars. Relaxed rows use the UMA relaxation procedure in Appendix~\ref{app:relax}.}
\end{table}

CG-OMatG matches NACJAF before relaxation (8/15 at $1.86$\,\AA) and after relaxation (11/15 at $0.37$\,\AA), and XAFPAY after relaxation (8/15 at $1.51$\,\AA); no method solves XAFQIH.

\paragraph{OXtal comparison on a training-disjoint OMC subset} OXtal's shipped checkpoint was trained on other members of the OMC test set used in Table~\ref{tab:omc128_results}, so that comparison is not apples-to-apples. We therefore selected all structures in our held-out 1,000-structure OMC test set that were absent from both OXtal's and CG-OMatG's training data, leaving 37 structures that were processed with OXtal's released routine. Table~\ref{tab:omc37_results} reports the resulting comparison.

\begin{table}[H]
  \centering
  \scriptsize
  \caption{CCDC packing-similarity metrics on the 37 OMC targets absent from both models' training data. Values are mean $\pm$ SEM over ten $K=30$ blocks.}
  \label{tab:omc37_results}
  \resizebox{\linewidth}{!}{%
  \begin{tabular}{@{}lccccc@{}}
    \toprule
    Method & Solved & Collisions allowed & Packing match & Per draw & Clash \\
    \midrule
    OXtal & $\boldsymbol{0.20\pm0.02}$ & $\boldsymbol{0.20\pm0.02}$ & $\boldsymbol{0.50\pm0.02}$ & $\boldsymbol{0.10\pm0.00}$ & $\boldsymbol{0.00\pm0.00}$ \\
    CG-OMatG & $0.05\pm0.00$ & $0.08\pm0.00$ & $0.39\pm0.01$ & $0.05\pm0.00$ & $0.43\pm0.00$ \\
    CG-OMatG-IRL & $0.06\pm0.00$ & $0.10\pm0.00$ & $0.49\pm0.01$ & $0.06\pm0.00$ & $0.25\pm0.00$ \\
    \bottomrule
  \end{tabular}}
\end{table}

OXtal has the higher solved rate on this subset, while CG-OMatG-IRL nearly closes the packing-match gap and reduces clashes relative to CG-OMatG. This does not invalidate CG-OMatG: a true apples-to-apples comparison remains difficult because of differing data splits and benchmarking pipelines. A fair comparison would require fully retraining OXtal on our split, which is currently impossible without a valid released training setup.

\begin{figure}[H]
    \centering
    \includegraphics[width=0.98\textwidth]{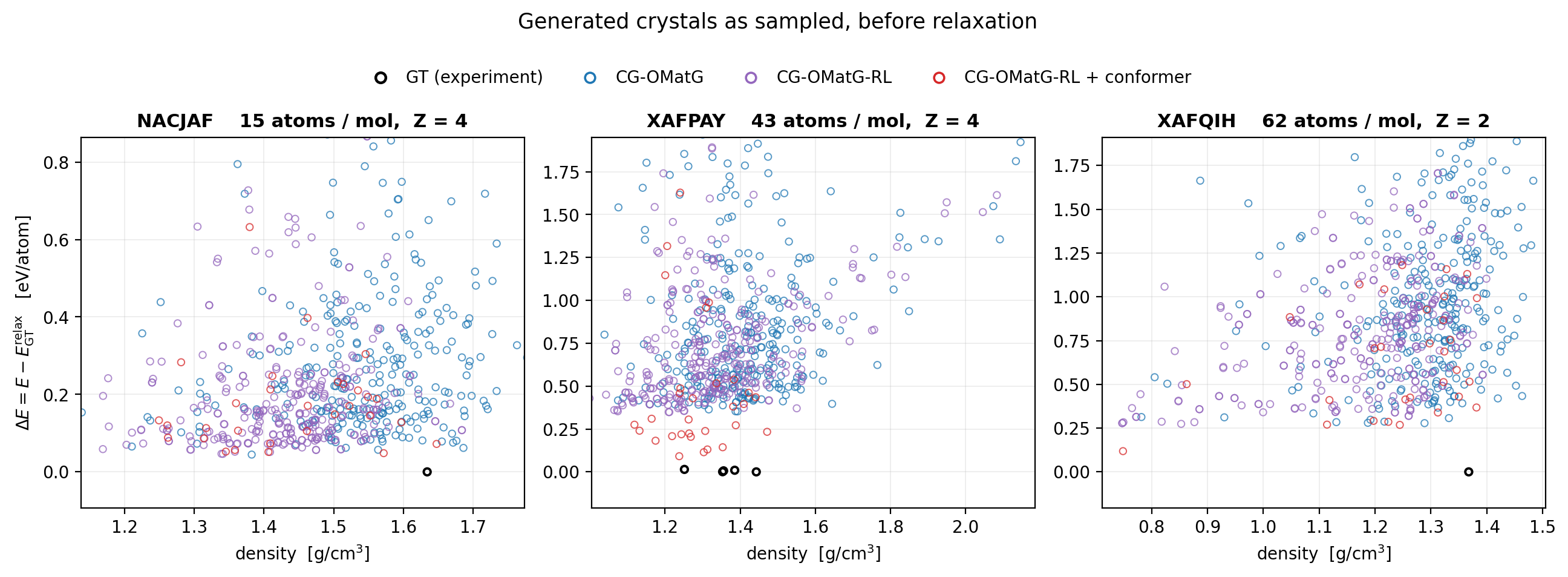}\\[0.5em]
    \includegraphics[width=0.98\textwidth]{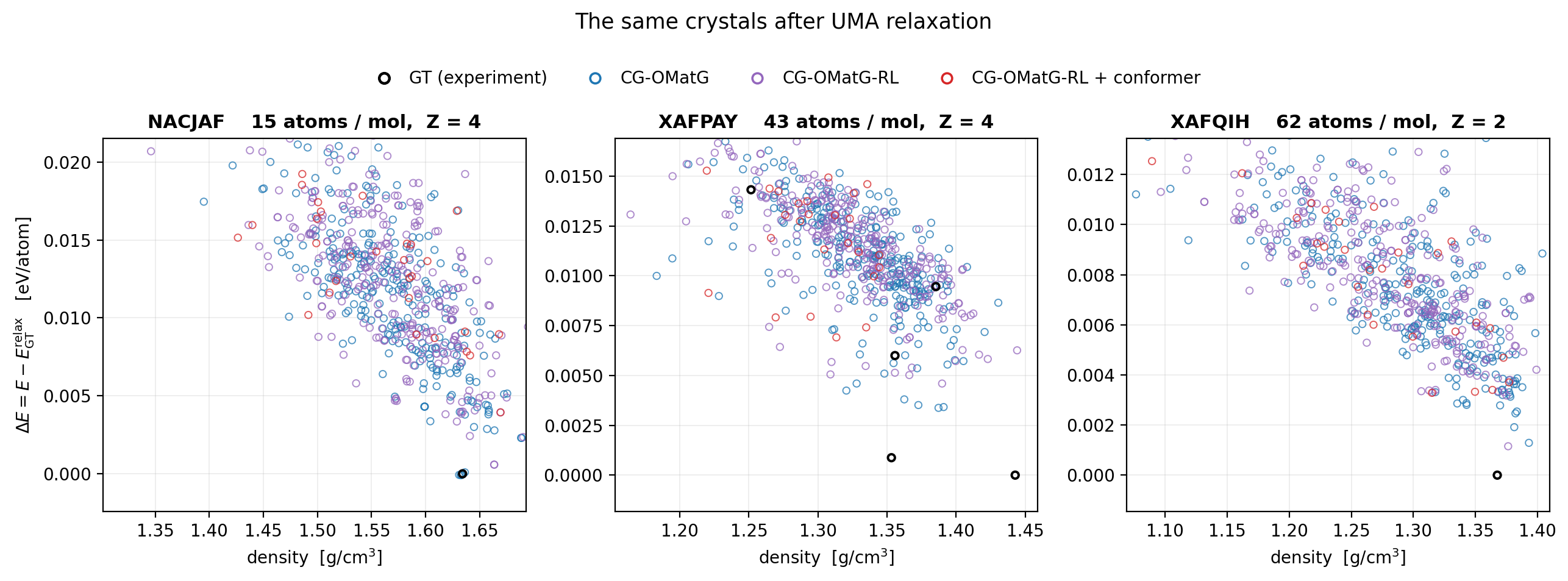}
    \caption{Energy versus density for generated sixth CSD blind-test structures before relaxation (top) and after UMA relaxation (bottom). Energies are reported relative to the UMA-relaxed experimental target. Relaxation sharpens the energy--density distributions for NACJAF, XAFPAY, and XAFQIH; for all three targets, the best CG-OMatG sample lies within $0.01$\,eV/atom of the lowest-energy relaxed ground truth with low density error.}
    \label{fig:energy_density_bt6}
\end{figure}

\paragraph{Energy and density before and after relaxation} Figure~\ref{fig:energy_density_bt6} compares the sampled structures with the same structures after UMA relaxation.

DFT would be required to make claims about small energy differences; the UMA evaluations here are intended to evaluate the methodology.

\begin{figure}[H]
    \centering
    \includegraphics[width=0.98\linewidth]{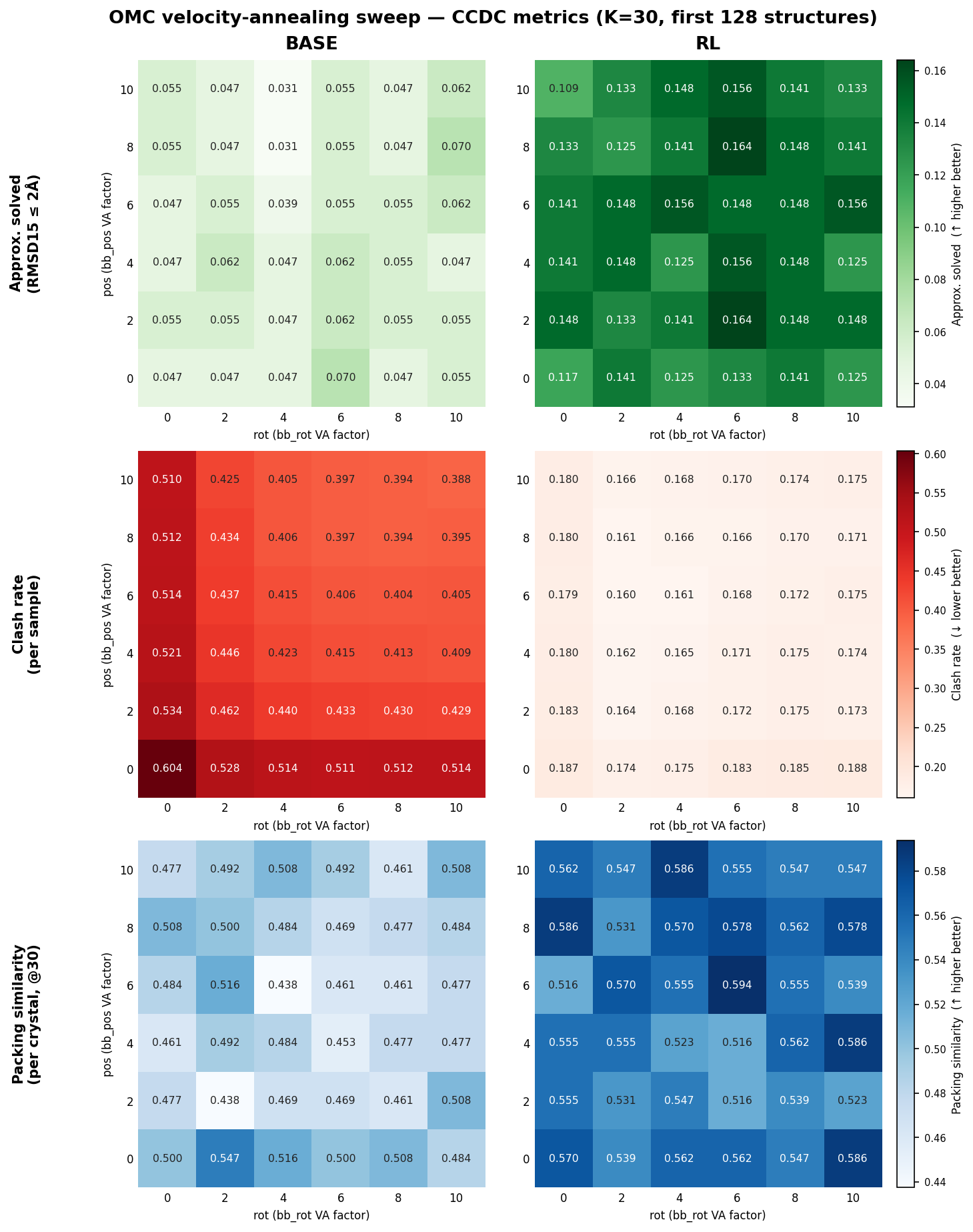}
    \caption{Full OMC128 velocity-annealing sweep for the base and reinforced models at $K=30$. Rows vary positional annealing and columns vary rotational annealing. Shown are solved rate, clash rate per draw, and target-level packing similarity.}
    \label{fig:annealing_heatmaps}
\end{figure}

\paragraph{Velocity annealing} We swept the positional and rotational velocity-annealing parameters over $s'_{\mathrm{pos}},s'_{\mathrm{rot}}\in\{0,2,4,6,8,10\}$ using $s(t)=1+s't$. Figure~\ref{fig:annealing_heatmaps} reports the full sweep.

\paragraph{Reward choice and circularity} The use of UMA in our reinforcement learning pipeline is partly circular, since OMC25-MCF was relaxed with UMA and UMA is also used to define the reward. To test how our results hinge on UMA as a reward, we reinforced the same pretrained checkpoint using Orb instead. The choice of Orb versus UMA as a reward, in this setting, does not markedly affect the robustness of the CG-OMatG-IRL strategy, as shown in Table~\ref{tab:omc128_results}. Most energy-function rewards may be expected to improve performance, especially on OMC. However, the performance gap between reinforced models on OMC and CSD data stems less from UMA's reward--data alignment with OMC and more from the difficulty of modeling experimental CSD data using an energy reward. The most reliable reward and assessment signal would be DFT, which is not scalable for reinforcement learning. Foundational MLIPs such as UMA, MACE, SevenNet, or Orb are reasonable proxies for evaluating the quality of proposal structures.

More broadly, using the same model for relaxation and reward does not invalidate the experiment. Crystal structure generation is a rare-event sampling problem because low-energy basins occupy only a small part of configuration space. Alignment with RL is intended to shift the proposal distribution toward these regions. Our experiment asks whether alignment reduces the number of proposals needed to recover held-out reference minima. This is an important practical goal of computational crystal structure prediction and serves as a key step toward a practical tool. Ideally, CSP models should also reproduce experimental structures; that is, however, not the specific question addressed in this work. Our aim is to determine whether alignment can make the sampling of relevant low-energy structures more efficient, which leaves the accuracy of these minima to those fitting MLIPs.

\begin{figure}[H]
    \centering
    \includegraphics[width=1\linewidth]{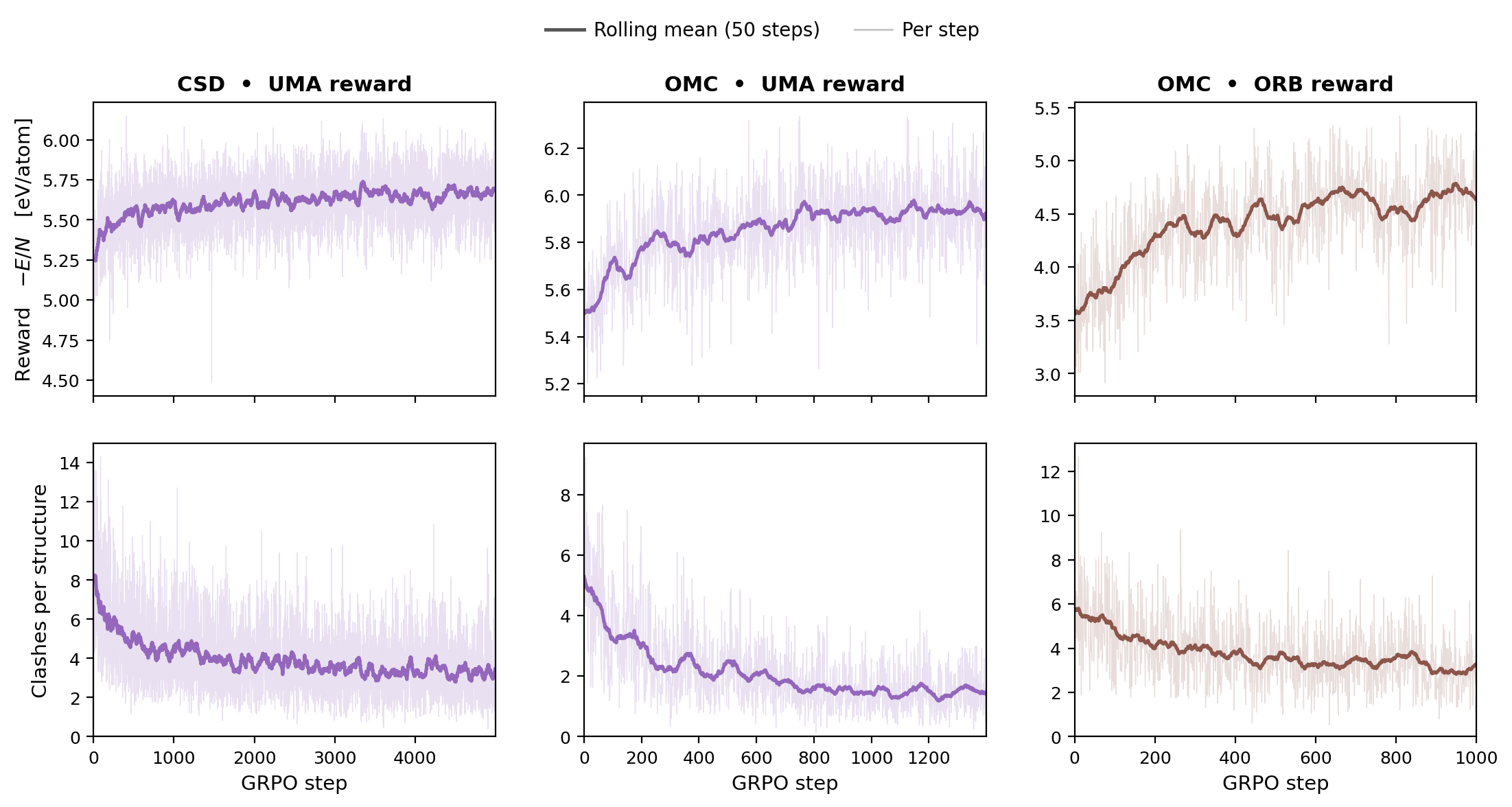}
    \caption{Reinforcement-learning trajectories for CSD with a UMA reward (left), OMC with a UMA reward (center), and OMC with an Orb reward (right). Top: per-step reward $-E/N$ and its 50-step rolling mean. Bottom: clashes per generated structure. Reward increases and clashes decrease across all three runs.}
    \label{fig:rl_reward_curves}
\end{figure}

\paragraph{Generated conformer inputs} In practice, the conformer is not known \emph{a priori}. To mimic this setting, we generated 500 ETKDGv3 conformers per molecule with RDKit, optimized and ranked them in vacuum using MMFF94s, and clustered the final geometries at a $0.50$\,\AA\ heavy-atom RMSD threshold. We retained the lowest-energy representative from each cluster and relaxed it with UMA. Table~\ref{tab:conformer_recovery} reports the closest recovered conformers; inference then used the top ten generated conformers.

\begin{table}[H]
  \centering
  \small
  \caption{Recovery of experimental blind-test conformers from ETKDGv3/MMFF94s sampling.}
  \label{tab:conformer_recovery}
  \resizebox{\linewidth}{!}{%
  \begin{tabular}{@{}lcccc@{}}
    \toprule
    CSD ID & Rotatable bonds & Energy rank & RMSD after MMFF94s & RMSD after UMA \\
    \midrule
    XAFQIH & 5 & 10 & $0.571$\,\AA & $0.648$\,\AA \\
    XAFPAY & 6 & 24 & $0.401$\,\AA & $0.348$\,\AA \\
    NACJAF & 0 & 1 & $0.057$\,\AA & $0.040$\,\AA \\
    \bottomrule
  \end{tabular}}
\end{table}

\begin{table}[H]
  \centering
  \scriptsize
  \caption{Blind-test metrics using generated conformer inputs. Each row has one $K=30$ block and therefore no error bars.}
  \label{tab:conformer_bt6}
  \resizebox{\linewidth}{!}{%
  \begin{tabular}{@{}llccccc@{}}
    \toprule
    Target & Pipeline & Solved & Collisions allowed & Packing match & Per draw & Clash \\
    \midrule
    NACJAF & IRL + conformer & \textbf{0.00} & \textbf{0.00} & \textbf{1.00} & \textbf{0.03} & 0.07 \\
           & IRL + conformer (relaxed) & \textbf{0.00} & \textbf{0.00} & \textbf{1.00} & \textbf{0.03} & \textbf{0.00} \\
    XAFPAY & IRL + conformer & \textbf{0.00} & \textbf{0.00} & \textbf{0.00} & \textbf{0.00} & 0.37 \\
           & IRL + conformer (relaxed) & \textbf{0.00} & \textbf{0.00} & \textbf{0.00} & \textbf{0.00} & \textbf{0.00} \\
    XAFQIH & IRL + conformer & \textbf{0.00} & \textbf{0.00} & \textbf{0.00} & \textbf{0.00} & 0.53 \\
           & IRL + conformer (relaxed) & \textbf{0.00} & \textbf{0.00} & \textbf{0.00} & \textbf{0.00} & \textbf{0.00} \\
    \bottomrule
  \end{tabular}}
\end{table}

The inexpensive conformer sampling produces candidate ensembles containing conformers close to the experimentally observed structures, but these inputs do not yield an additional solved blind-test target.

\paragraph{Polymorph diversity} To quantify diversity after RL post-training, we construct a graph over each set of $K=30$ generated crystals. Nodes $i$ and $j$ are connected when COMPACK aligns at least eight of fifteen molecules with $\mathrm{RMSD}_{N}<2.0$\,\AA. The number of connected components $C$ is the first diversity measure. Because $C$ can overstate diversity when many components are singletons, we also report the effective number of components $\mathrm{eff}=1/\sum_k p_k^2$, where $p_k=s_k/30$, and the dominant-mode fraction $\max_k p_k$. Results in Table~\ref{tab:diversity} are averaged over ten independent sets of 30 samples per target and reported as mean $\pm$ standard error.

\begin{table}[H]
  \centering
  \small
  \caption{Diversity of generated blind-test structures under COMPACK component clustering.}
  \label{tab:diversity}
  \resizebox{\textwidth}{!}{%
  \begin{tabular}{@{}llccc@{}}
    \toprule
    Target & Method & Distinct components $C$ & Effective components $\mathrm{eff}$ & Dominant-mode fraction $\max_k p_k$ \\
    \midrule
    NACJAF & CG-OMatG & $\boldsymbol{23.0\pm1.0}$ & $\boldsymbol{18.4\pm1.2}$ & $\boldsymbol{0.13\pm0.01}$ \\
    NACJAF & CG-OMatG-IRL & $21.6\pm1.2$ & $15.1\pm1.5$ & $0.17\pm0.01$ \\
    XAFPAY & CG-OMatG & $\boldsymbol{27.5\pm0.6}$ & $\boldsymbol{25.2\pm1.2}$ & $\boldsymbol{0.08\pm0.01}$ \\
    XAFPAY & CG-OMatG-IRL & $26.9\pm0.4$ & $24.4\pm0.8$ & $\boldsymbol{0.08\pm0.01}$ \\
    XAFQIH & CG-OMatG & $\boldsymbol{27.1\pm0.8}$ & $\boldsymbol{23.9\pm1.9}$ & $\boldsymbol{0.10\pm0.02}$ \\
    XAFQIH & CG-OMatG-IRL & $21.5\pm1.0$ & $12.9\pm1.7$ & $0.23\pm0.04$ \\
    \bottomrule
  \end{tabular}}
\end{table}

While CG-OMatG-IRL does exhibit a reduced diversity score, we do not believe that this number is indicative of mode collapse but rather suggests more refined inference with respect to the UMA energy landscape.

\section{Loss Curves}

\begin{figure}[H]
    \centering
    \includegraphics[width=0.85\linewidth]{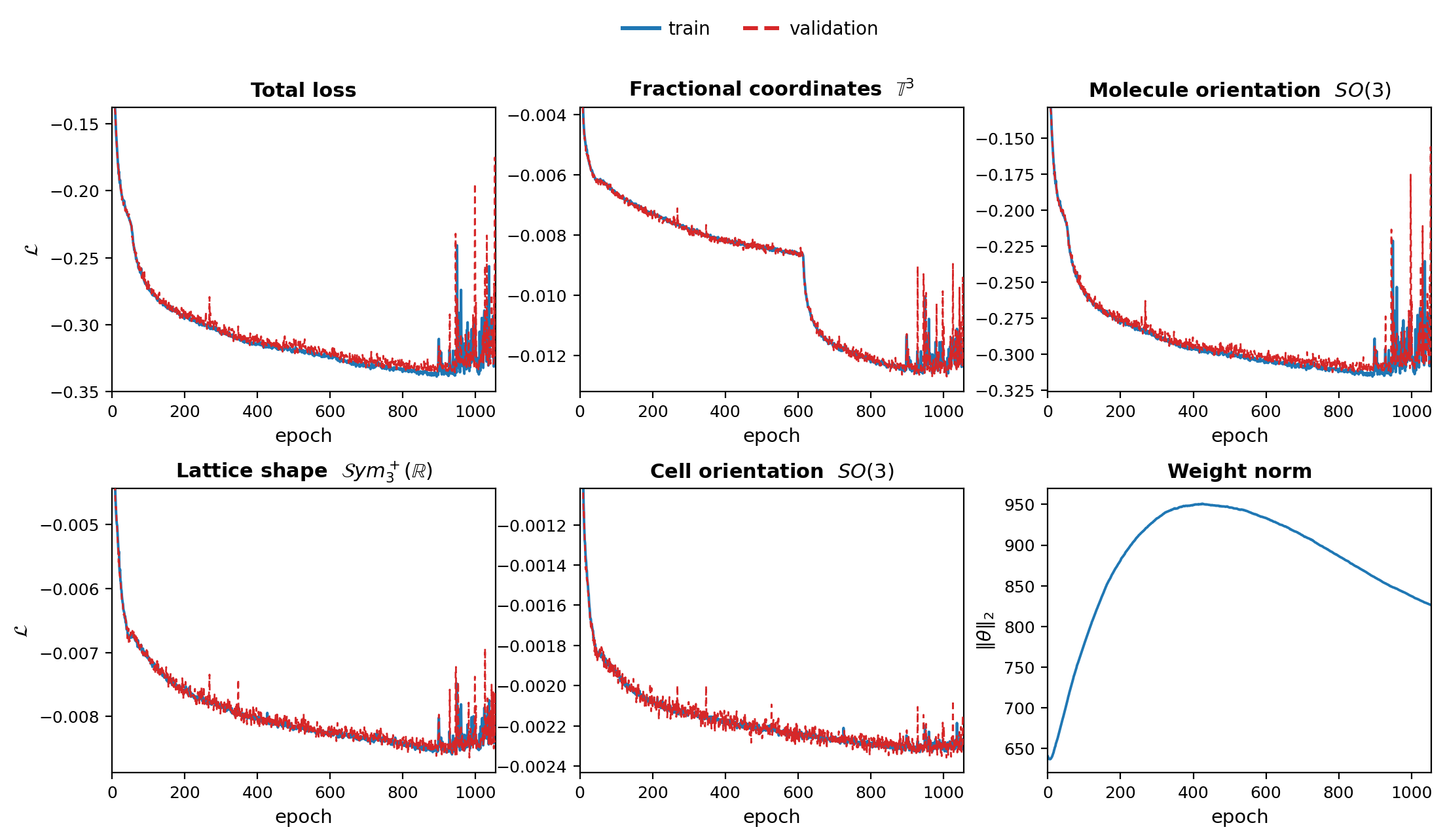}
    \caption{CG-OMatG training and validation losses on the CSD database, decomposed by manifold component: lattice shape $\mathrm{Sym}_3^+$, cell and molecule orientations on $SO(3)$, and fractional coordinates on $\mathbb{T}^3$. Weight norm is the global L2 norm of all trainable parameters. Note that the constant term is dropped from this loss, so it can be below zero, unlike the usual normalizing-flow presentation.}
\end{figure}

\begin{figure}[H]
    \centering
    \includegraphics[width=0.85\linewidth]{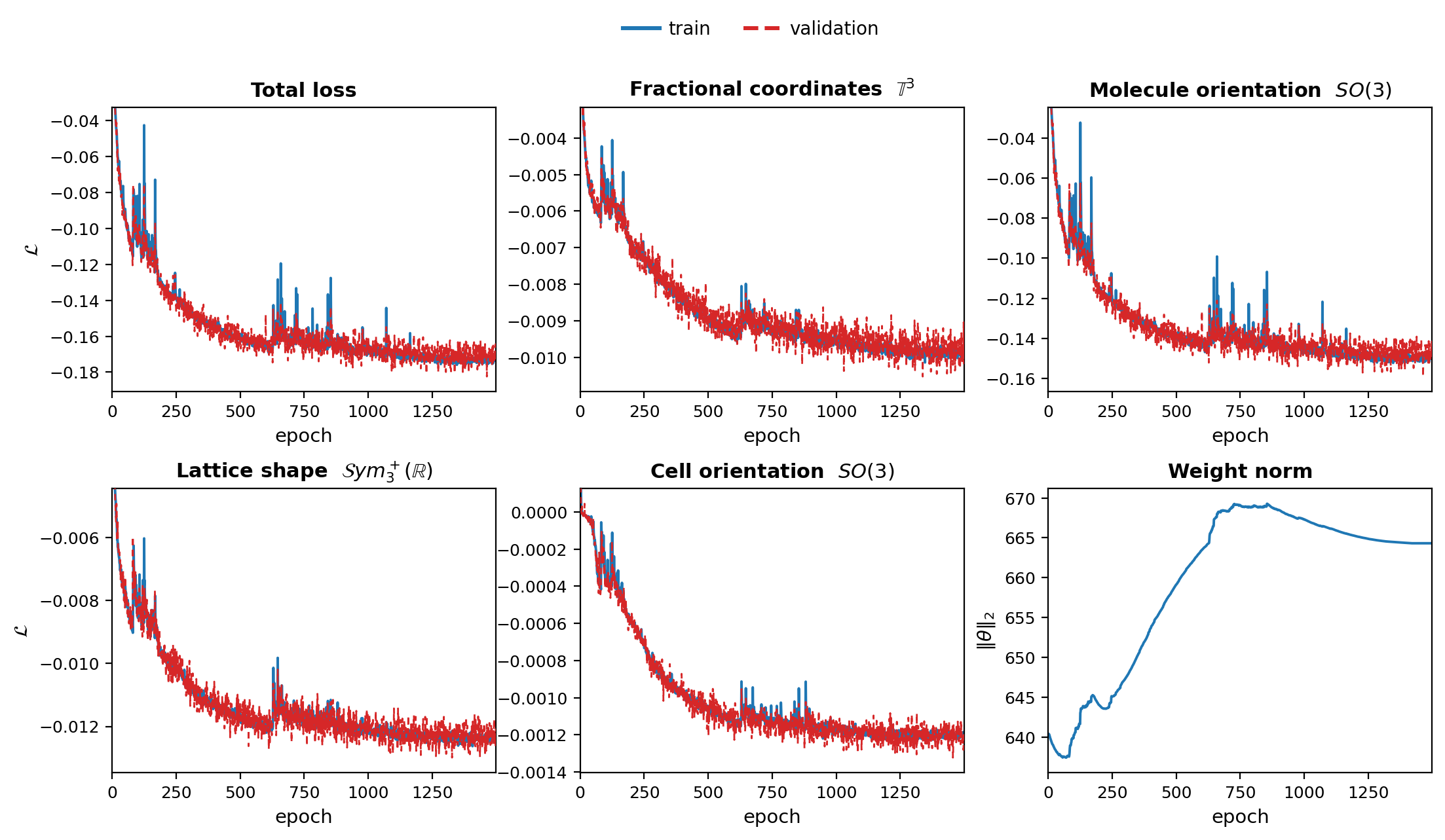}
    \caption{CG-OMatG training and validation losses on the OMC database, decomposed by manifold component: lattice shape $\mathrm{Sym}_3^+$, cell and molecule orientations on $SO(3)$, and fractional coordinates on $\mathbb{T}^3$. Weight norm is the global L2 norm of all trainable parameters. Note that the constant term is dropped from this loss, so it can be below zero, unlike the usual normalizing-flow presentation. Occasional jumps in the loss curves reflect checkpoint restarts after improper GPU resource allocation on the cluster.}
  \label{fig:training_curves_omc}
\end{figure}

\section{Data Availability, Preprocessing, and Resources}\label{app:data_processing}

\paragraph{OMC25-MCF} This OMC subset is processed by \citet{zeng_molcrystalflow_2026}, as described in Section~\ref{sec:data_mainpaper}. Specifically, all cocrystals are filtered out, leaving only homomolecular crystals. Subsequently, for each crystal family, the \texttt{uma-s-1p1} MLIP \citep{wood_uma_2026} is used to retain the crystal polymorph with the lowest energy per conformer. After filtering, the dataset is reduced to $46,120$ molecular crystal structures.

\paragraph{CSD} The CSD dataset is processed similarly to \citep{jin_oxtal_2025, subramanian_packflow_2026}. First, we filter out any crystal unit cells containing more than $250$ heavy atoms. Then, we prescribe that no member of the CSD blind test crystal families may be present in the training data and that the SMILES are indeed valid SMILES strings using RDKit. We ensure that the crystal unit cells possess 3-D coordinates and an $R$-factor $< 0.9$. The crystal must have a space group symbol. We resolve disorder by selecting the disorder group with the highest occupancy. Lastly, we split the data into training, validation, and test sets such that crystal polymorphs of the same family belong to the data split. To resolve degeneracy, we compare polymorphs and ensure that the RMSD between them does not fall below $0.25$\,\AA; if it does, we retain the polymorph with the lowest $R$-factor.

\paragraph{Resources Used} All training and inference were carried out on NVIDIA A100 GPUs (80\,GB HBM2e) on a shared SLURM cluster. Each training run used $4 \!\times\! \text{A100}$ in a single node under PyTorch Lightning DDP at FP32 precision. For evaluation, generation is carried out on the same hardware. Downstream CCDC packing-similarity and relaxation-based metrics run on CPU-only nodes (16 cores, 60\,GB RAM, $\le\!3$\,h walltime per dataset), parallelized across structures with a single L40 GPU for UMA energy calculations. 
\end{document}